\documentclass[10pt,twocolumn,letterpaper]{article}

\usepackage[pagenumbers]{cvpr} 

\definecolor{cvprblue}{rgb}{0.21,0.49,0.74}
\usepackage[pagebackref,breaklinks,colorlinks,allcolors=cvprblue]{hyperref}
\usepackage[table]{xcolor}
\usepackage{multirow}

\newcommand{\ours}{Ours}
\def\paperID{258} 
\def\confName{3DV\xspace}
\def\confYear{2027\xspace}

\title{PhGS: Post‑Hoc Pruning and Refinement of\\Single‑View Feed‑Forward 3D Gaussian Reconstructions}

\author{
    Rinto Yagawa\textsuperscript{1}\quad
    Han Cheng\textsuperscript{2}\quad
    Dieter Schmalstieg\textsuperscript{2}\quad
    Hideo Saito\textsuperscript{1}\quad
    Shohei Mori\textsuperscript{2}\\
    \textsuperscript{1}Keio University\quad
    \textsuperscript{2}University of Stuttgart
}

\begin{document}

\twocolumn[{%
\renewcommand\twocolumn[1][]{#1}%
\maketitle
\includegraphics[width=\textwidth]{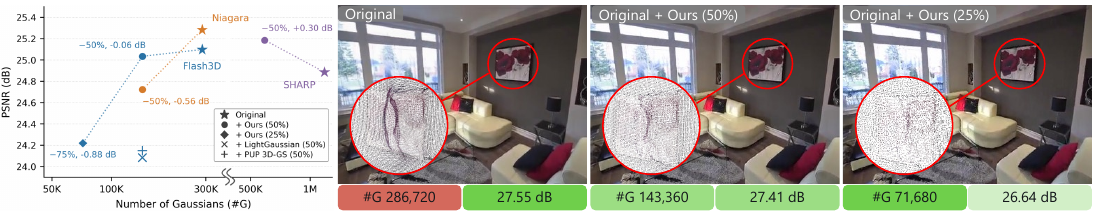}
\vspace{-2.2em}
\captionof{figure}{
Our post-hoc compaction method seamlessly integrates with any single-view feed-forward 3D Gaussian Splatting model. Pre-trained baselines such as Flash3D~\cite{szymanowiczFlash3dFeedforwardGeneralisable2025}, Niagara~\cite{wuNiagaraNormalIntegratedGeometric2025a} and SHARP~\cite{meschederSharpMonocularView2026} predict dense Gaussians; our method prunes redundant primitives and refines the remaining ones, achieving arbitrary memory reduction while largely preserving or even improving the novel-view rendering fidelity. Left: performance across backbones, with LightGaussian~\cite{fan2024lightgaussian} and PUP 3D-GS~\cite{hanson2025pup} applied to Flash3D~\cite{szymanowiczFlash3dFeedforwardGeneralisable2025} as post-hoc pruning methods; right: rendering quality with Flash3D~\cite{szymanowiczFlash3dFeedforwardGeneralisable2025}. The insets show the Gaussian center points.
}
\vspace{1em}
\label{fig:teaser}
}]

\begin{abstract}
Recent single-view feed-forward 3D Gaussian Splatting (3DGS) generation predicts a fixed number of Gaussians per camera ray, introducing severe spatial redundancy. Most existing compaction strategies target multi-view setups to exploit cross-view consistency and are incompatible with single-image models. Instead of retraining the base feed-forward network to directly output compact representations, our insight is to keep the base models frozen and apply post-hoc pruning and recurrent refinement to the generated Gaussians. Consequently, we propose a backbone-agnostic compaction pipeline for single-view feed-forward 3DGS that couples an importance-score-based pruning mechanism with a trainable, lightweight recurrent refinement module, which iteratively updates the surviving primitives to restore image quality.
Our results demonstrate seamless integration with existing baselines while preserving novel-view rendering fidelity and achieving high memory reduction. 
Furthermore, our method supports flexible inference-time keep ratios for application needs.
\end{abstract}
\section{Introduction}
\label{sec:intro}
Reconstructing a scene with 3D Gaussian Splatting (3DGS) \cite{kerbl3dGaussianSplatting2023} from a single feed-forward pass is a core problem in modern computer vision.
In multi-view setups \cite{charatanPixelsplat3dGaussian2024, chenMVSplatEfficient3D2025c, chenMvsplat360Feedforward3602024, xuDepthsplatConnectingGaussian2025, zhangGSLRMLargeReconstruction2025a, yeNoPoseNo2025, yeYoNoSplatYouOnly2025b}, networks typically reconstruct a scene by assigning exactly one Gaussian primitive to each pixel across all views. Single-view methods  \cite{szymanowiczSplatterImageUltrafast2024, szymanowiczFlash3dFeedforwardGeneralisable2025, wuNiagaraNormalIntegratedGeometric2025a, meschederSharpMonocularView2026, xieHiGaussianHierarchicalGaussians2025a, diaoMultiLayerGaussianSplatting2025b}, however, cannot rely on such multi-view constraints to reason occluded content along each ray from an input view, and instead, predict $K~(> 1)$ Gaussians per camera ray.
The $K$-Gaussians-per-ray strategy produces significant spatial redundancy that is uninformative for uniform regions or single-surface rays, thereby inflating memory consumption.


Despite this inherent issue, existing research on compact feed-forward 3DGS \cite{smartLearningCompact3D, jiangAnySplatFeedforward3D2025b, liuWorldMirrorUniversal3D2025, wangVolSplatRethinkingFeedForward2026b, wangFreesplatGeneralizable3d2024, parkEcoSplatEfficiencycontrollableFeedforward2026a, kimF4SplatFeedForwardPredictive2026a, zhangSparseSplatApplicableFeedForward2026a, renTokenGSDecoupling3D2026, liTokenSplatTokenaligned3D2026, moreauGridDetectionPrimitives2026, itkinGlobalSplatEfficientFeedForward2026a, zhangAnchorSplatFeedForward3D2026a} focuses almost exclusively on the multi-view setting.
These approaches typically target geometric fusion to consolidate overlapping primitives across different viewpoints \cite{smartLearningCompact3D, jiangAnySplatFeedforward3D2025b, liuWorldMirrorUniversal3D2025, wangVolSplatRethinkingFeedForward2026b, wangFreesplatGeneralizable3d2024} or learn to allocate adaptive capacity \cite{parkEcoSplatEfficiencycontrollableFeedforward2026a, kimF4SplatFeedForwardPredictive2026a, zhangSparseSplatApplicableFeedForward2026a, renTokenGSDecoupling3D2026, liTokenSplatTokenaligned3D2026, moreauGridDetectionPrimitives2026, itkinGlobalSplatEfficientFeedForward2026a, zhangAnchorSplatFeedForward3D2026a}.
Because their sparsification mechanisms are learned jointly with multi-view backbones, they cannot be trivially adapted to the single-view domain. This gap motivates the development of a dedicated sparsification method for single-view feed-forward 3DGS.  

To this end, we introduce a cross-model sparsification framework customized for single-view feed-forward 3DGS.
To avoid modifying the internal mechanisms of base models, our method operates entirely on the output space. We first propose an importance-score-based pruning mechanism that directly assesses importance using generated Gaussian opacities and 2D image priors. To compensate for the pruned primitives, we employ a recurrent refinement module that directly updates the parameters of the surviving Gaussians.
By decoupling sparsification from the underlying network, our framework seamlessly integrates with diverse models. It requires only a lightweight refinement module trained on the outputs of a frozen base model.

To demonstrate the universality of our pipeline, we evaluate our framework on the RealEstate10K \cite{zhouStereoMagnificationLearning2018c} dataset across three state-of-the-art single-view feed-forward 3DGS architectures, namely Flash3D~\cite{szymanowiczFlash3dFeedforwardGeneralisable2025}, Niagara~\cite{wuNiagaraNormalIntegratedGeometric2025a}, and SHARP~\cite{meschederSharpMonocularView2026}.
As illustrated in \figurename ~\ref{fig:teaser}, our method successfully reduces the number of Gaussians to an arbitrary target count across
these distinct models while preserving the quality of novel view synthesis.
This backbone-agnostic design allows developers to flexibly select the base estimator that best fits their application needs, ranging from lightweight to efficient deployment.
Ultimately, because our framework is decoupled from the base network structure, it provides an adaptable solution for emerging single-view feed-forward 3DGS architectures.
Our primary contributions are as follows:
\begin{itemize}
    \item To the best of our knowledge, this is the first \textit{post-hoc pruning and refinement} method for single-view feed-forward 3DGS. By decoupling the sparsification from the base architecture, our pipeline seamlessly integrates with a variety of off-the-shelf Gaussian estimators to effectively resolve spatial redundancy.
    \item We introduce an effective pruning-and-refinement pipeline that stably supports a wide range of Gaussian keep ratios at inference time, enabling flexible trade-offs between memory and rendering quality.
    \item We extensively validate our framework on the RealEstate10K dataset \cite{zhouStereoMagnificationLearning2018c}. We demonstrate its universality by reducing the number of Gaussians to user-specified keep ratios across three distinct state-of-the-art architectures with minimal rendering degradation.
\end{itemize}



\section{Related Work}
\subsection{Feed-Forward 3D Gaussian Splatting}
Feed-forward 3D Gaussian Splatting instantly generates an explicit 3D representation from images through a single network forward pass, bypassing the expensive per-scene optimization of standard 3DGS \cite{kerbl3dGaussianSplatting2023}. 

Initially, multi-view approaches \cite{charatanPixelsplat3dGaussian2024, chenMVSplatEfficient3D2025c, chenMvsplat360Feedforward3602024, xuDepthsplatConnectingGaussian2025, zhangGSLRMLargeReconstruction2025a, yeNoPoseNo2025, yeYoNoSplatYouOnly2025b} achieved generalizable sparse-view synthesis by predicting exactly one Gaussian primitive per pixel. The network leverages cross-view consistency to establish coherent 3D structures.
In contrast, single-view feed-forward 3DGS methods \cite{szymanowiczSplatterImageUltrafast2024, szymanowiczFlash3dFeedforwardGeneralisable2025, wuNiagaraNormalIntegratedGeometric2025a, meschederSharpMonocularView2026, xieHiGaussianHierarchicalGaussians2025a, diaoMultiLayerGaussianSplatting2025b} lack multi-view consistency. They typically rely on monocular depth priors \cite{piccinelliUnidepthUniversalMonocular2024, bochkovskiyDepthProSharp2025} and deliberately allocate multiple Gaussians ($K$) along each camera ray to explicitly model occluded regions. For instance, Flash3D \cite{szymanowiczFlash3dFeedforwardGeneralisable2025} predicts two Gaussians per pixel, while subsequent multi-layer strategies \cite{diaoMultiLayerGaussianSplatting2025b, xieHiGaussianHierarchicalGaussians2025a} extend this by placing a denser set of $K$ primitives per pixel.

However, a fundamental limitation persists across all these architectures: forcing a fixed number of primitives ($1$ or $K$) uniformly across every pixel is inherently inefficient. 
Real-world scenes consist of large flat regions and detailed regions. Consequently, the rigid, pixel-aligned allocation creates severe spatial redundancy and memory waste, motivating our importance-score based pruning framework.

\begin{figure*}[t] 
  \centering
  \includegraphics[width=\textwidth]{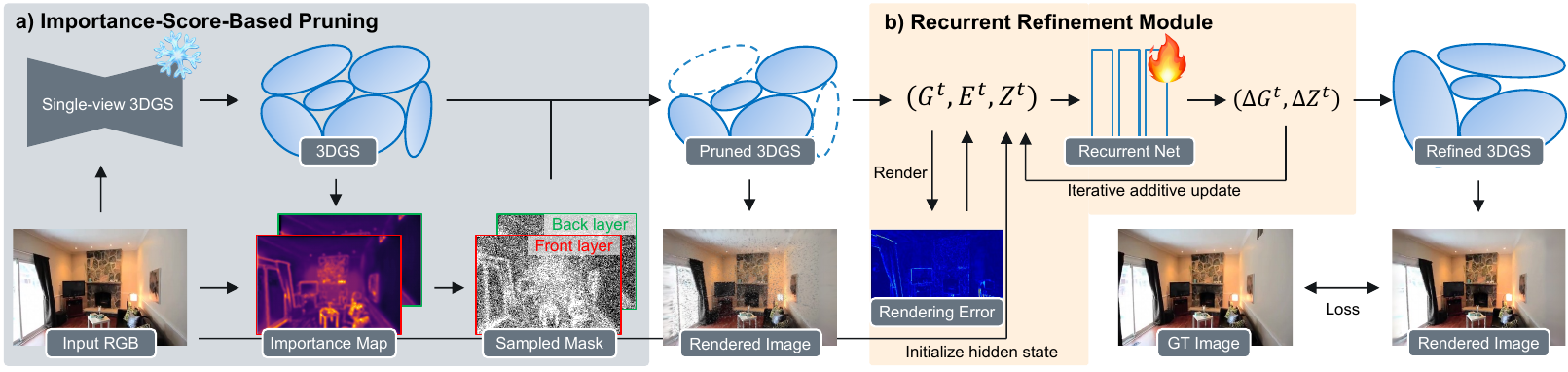}
  \caption{Overview of our cross-model post-hoc compaction pipeline. The pipeline operates in two phases. (a) We compute a spatial importance map and probabilistically sample a mask to prune redundant primitives (Sec.~\ref{sec:pruning}). (b) The surviving Gaussians undergo iterative updates to recover rendering quality (Sec.~\ref{sec:rnn}). At each step, the recurrent network predicts parameter updates ($\Delta G^t, \Delta Z^t$) from the current Gaussian parameters $G^t$, the hidden state $Z^t$, and the rendering error  feature $E^t$.}
  \label{fig:pipeline}
  \vspace{-10pt}
\end{figure*}

\subsection{Compact Feed-Forward 3DGS}
\label{sec: cff3dgs}
Recognizing the severe memory inefficiency of dense feed-forward predictions, recent work has explored methods that produce compact 3DGS representations. These approaches use cross-view geometric fusion or adaptive capacity allocation to reduce the number of Gaussians, predominantly in the multi-view setting.

The first theme focuses on geometric and structural fusion, using octree-based point merging \cite{smartLearningCompact3D}, differentiable voxelization \cite{jiangAnySplatFeedforward3D2025b, liuWorldMirrorUniversal3D2025, wangVolSplatRethinkingFeedForward2026b}, and pixel-wise triplet fusion \cite{wangFreesplatGeneralizable3d2024}. In multi-view feed-forward setups, a major source of redundancy arises from independent per-view predictions; the network often generates multiple, slightly misaligned Gaussians to represent the same 3D physical surface due to cross-view inconsistency. These fusion methods are especially effective at consolidating overlapping, inconsistent primitives. However, this form of redundancy is fundamentally absent in the single-image scenario, where primitives are already tightly organized along structured camera rays.

The second theme involves adaptive capacity allocation. Methods such as EcoSplat~\cite{parkEcoSplatEfficiencycontrollableFeedforward2026a}, F4Splat~\cite{kimF4SplatFeedForwardPredictive2026a}, and SparseSplat~\cite{zhangSparseSplatApplicableFeedForward2026a} modulate the local Gaussian density using learned densification scores or entropy-based sampling. Similarly, token-based approaches \cite{renTokenGSDecoupling3D2026, liTokenSplatTokenaligned3D2026, moreauGridDetectionPrimitives2026, itkinGlobalSplatEfficientFeedForward2026a} and anchor-aligned models like AnchorSplat~\cite{zhangAnchorSplatFeedForward3D2026a} replace spatial grids with a fixed number of latent scene tokens or geometric anchors that decode directly into 3DGS parameters. However, extending these dynamic allocation strategies to single-view reconstruction poses a fundamental challenge. Because these mechanisms are learned jointly with specific multi-view architectures \cite{wang2025vggt, xuDepthsplatConnectingGaussian2025}, they cannot be applied to an existing single-view predictor without retraining, and adapting them to a single input requires architecture-specific modifications, causing degradation of rendering quality.

To bypass these architectural dependencies, our work delivers a backbone-agnostic sparsification framework customized for single-view feed-forward 3DGS. Our approach operates directly on the output Gaussians to obtain a compact representation without requiring structural modifications to the underlying backbone.


\subsection{Inference-time Refinement}
Recent feed-forward 3D Gaussian Splatting methods demonstrate impressive reconstruction quality with efficient inference \cite{charatanPixelsplat3dGaussian2024, chenMVSplatEfficient3D2025c, chenMvsplat360Feedforward3602024, xuDepthsplatConnectingGaussian2025, zhangGSLRMLargeReconstruction2025a, yeNoPoseNo2025, yeYoNoSplatYouOnly2025b}. However, these one-shot prediction pipelines are fundamentally bounded by network capacity, particularly when constrained to output highly compact representations. To overcome this limitation, several works augment feed-forward 3DGS with inference-time refinement.

Existing approaches refine the initial Gaussians either by incorporating test-time gradient-based optimization \cite{liuDiff3RFeedforward3D2026a} or by applying gradient-free iterative updates \cite{xuReSplatLearningRecurrent2026a, chenGIFSplatGenerativePriorGuided2026a}. These methods progressively improve reconstruction quality through additional inference-time computation. For instance, ReSplat \cite{xuReSplatLearningRecurrent2026a} predicts Gaussians in a 16x subsampled space and then applies refinement. While the compression strategy is efficient, the method is tightly coupled to the base feed-forward 3DGS model \cite{xuDepthsplatConnectingGaussian2025}, which restricts its applicability.

In contrast, our method introduces a lightweight recurrent refinement module that operates directly on the outputs of a frozen base model. By training this module, our method provides model-agnostic inference-time refinement for arbitrary single-view feed-forward 3DGS models.

\section{Method}
Given a single RGB image $I$, our goal is to reconstruct a 3D scene using a compact set of $m$ 3D Gaussian primitives ($m \ll H \times W \times K$). Here, $H \times W$ denotes the pixel grid (which may differ from the input image resolution) at which the underlying feed-forward backbone predicts Gaussians, and $K$ is the number of Gaussians per pixel. Recent single-view feed-forward 3DGS methods \cite{szymanowiczFlash3dFeedforwardGeneralisable2025, wuNiagaraNormalIntegratedGeometric2025a, meschederSharpMonocularView2026} predict $K=2$ Gaussians for each pixel to model occluded surfaces explicitly. As illustrated in \figurename ~\ref{fig:pipeline}, our framework operates on the output of such estimators. The importance-score-based pruning mechanism (Sec.~\ref{sec:pruning}) reduces the initial $H \times W \times K$ Gaussians to a smaller set of $m$ primitives. Subsequently, the recurrent refinement module (Sec.~\ref{sec:rnn}) iteratively updates the parameters of the surviving $m$ Gaussians directly.

\subsection{Importance-Score-Based Pruning}
\label{sec:pruning}
\noindent\textbf{Importance Calculation.} 
To reduce the dense representation $\mathcal{G}_{\mathrm{dense}} \in \mathbb{R}^{H \times W \times K \times C}$, where $C$ denotes the dimensionality of the per-Gaussian parameters, we introduce a probabilistic pruning mechanism that retains a fixed number of Gaussians via weighted sampling. 
Inspired by recent work demonstrating that edge information is informative for Gaussian primitive allocation \cite{meulemanOntheflyReconstructionLargeScale2025a}, we define an importance score $S_{k,i}$ for the $k$-th Gaussian layer at pixel $i$ by combining an edge prior with predicted opacity. 
\begin{equation}
S_{k,i} = \lambda_{\mathrm{edge}} E_i + \lambda_{\mathrm{op}} \alpha_{k,i},
\end{equation} 
where $k \in {1,\ldots,K}$ indexes the $K$ Gaussians predicted per pixel, $E_i$ denotes the edge response at pixel $i$ computed from the input image $I$ using a Laplacian operator, and $\alpha_{k,i}$ denotes the opacity of the $k$-th Gaussian at pixel $i$. 
This score encourages selecting high-opacity primitives while preserving structural boundaries. We use the same score formulation for all $K$ layers because, in single-view methods, Gaussians predicted on the deeper layers are not tied to a fixed semantic role; instead, they may support foreground geometry or represent distinct background regions depending on the pixel. 
\begin{figure}[t] 
  \centering
  \includegraphics[width=\columnwidth]{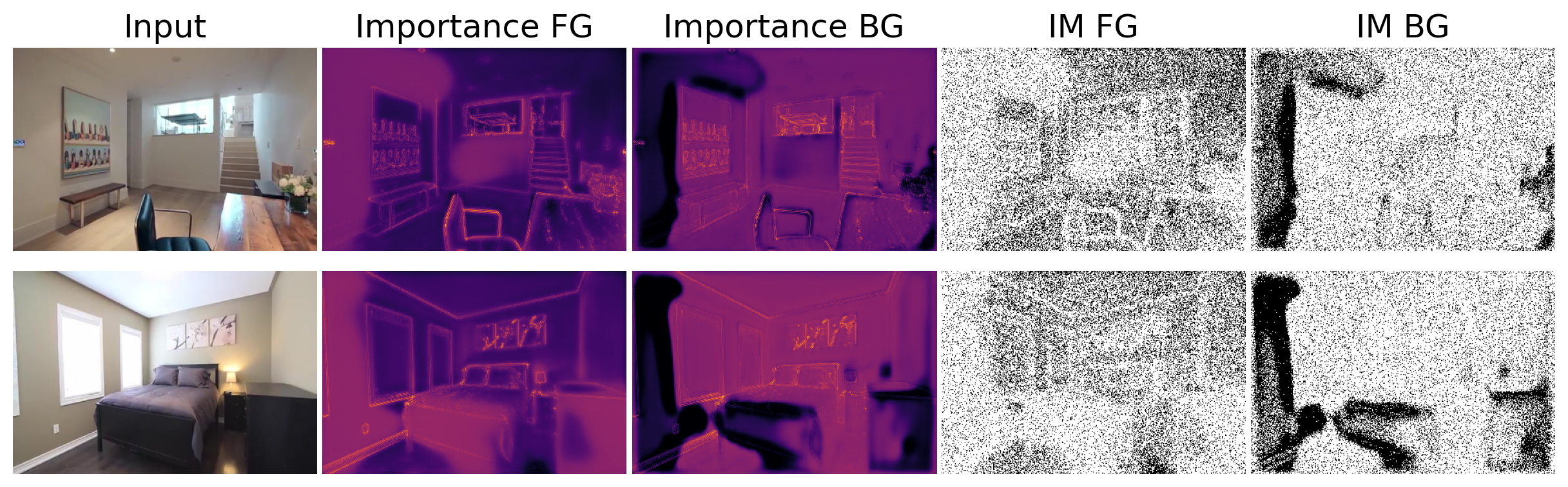}
  \caption{Visualization of the calculated importance map and sampled Importance Mask (IM) from the Flash3D backbone for both the front (FG) and back (BG) layers ($r=0.5$). Maps and masks are shown over the pixel grid of the source image.}
  \label{fig:mask_IM}
  \vspace{-10pt}
\end{figure}

\vspace{0.5em}
\noindent\textbf{Sampling.}
\label{sec:sampling}
In our method, we sample a two-layer binary mask $\mathcal{M} \in \{0, 1\}^{H \times W \times K}$, where each entry indicates whether the corresponding Gaussian is retained.
Given a target keep ratio $r \in (0, 1]$, we sample entries $m:= \lfloor r \times H \times W \times K \rfloor$ without replacement with probabilities proportional to the importance scores $S$, prioritizing Gaussians anchored at source-image pixels. Rather than retaining a fixed proportion of pixels in each layer, we sample according to the distribution of importance scores across both layers to generate a binary mask $\mathcal{M}$. We refer to the resulting mask as the \textbf{Importance Mask (IM)} (\figurename ~\ref{fig:mask_IM}). Finally, we retain the Gaussians corresponding to the pixels selected by this IM and prune all remaining Gaussians. Ultimately, we obtain a compact set $\mathcal{G}_{\mathrm{compact}} \in \mathbb{R}^{m \times C}$.


\subsection{Recurrent Refinement Module}
\label{sec:rnn}
While our pruning method efficiently removes spatial redundancy, naively discarding primitives inevitably leaves gaps in the 3D representation and degrades novel-view synthesis. To restore visual fidelity, we attach a lightweight recurrent refinement module that directly updates the surviving Gaussian parameters over a few iterations.

\vspace{0.5em}
\noindent\textbf{Iterative Additive Update.}
At iteration $t$, the module ingests three signals: the current Gaussian parameters $G^t$, a per-Gaussian hidden state $Z^t$, and a per-Gaussian error feature $E^t$ (defined below). We initialize the hidden state as $Z^0 = \pi(\phi(I))$, where the source image $I$ is passed through a frozen VGG encoder $\phi$, and $\pi$ is a per-Gaussian gather operation. The network predicts additive updates
\begin{equation}
    (\Delta G^t, \Delta Z^t) = f_\theta(G^t, Z^t, E^t), 
\end{equation}
which are applied as
\begin{equation}
\label{eq:update_g_z}
    G^{t+1} = G^t + L \odot \tanh(\Delta G^t),  Z^{t+1} = Z^t + \Delta Z^t. 
\end{equation}
Here, $L \in \mathbb{R}^C$ is a predefined scaling vector bounding the maximum allowable parameter updates.

Inspired by ReSplat \cite{xuReSplatLearningRecurrent2026a}, we utilize a recurrent update mechanism. However, to ensure scalability for the dense Gaussian counts typical of single-view feed-forward 3DGS estimators (e.g., SHARP~\cite{meschederSharpMonocularView2026} outputs $>1$M Gaussians), we strictly avoid computationally expensive spatial operations such as kNN or global attention. Instead, we use a lightweight two-layer MLP $f_\theta$ applied independently to each Gaussian to update the parameters. 
We concatenate each Gaussian's local feature with a global-context feature obtained by average pooling over all Gaussians, then feed the results to $f_\theta$. This provides scene-level coordination via a simple shared vector that scales linearly with the number of Gaussians.

\vspace{0.5em}
\noindent\textbf{Per-Gaussian Error Feature.}
At each iteration, the surviving Gaussians are rasterized at the source view, producing a rendered image $\hat{I}^t$. We compute the residual against the input image in both pixel and perceptual feature spaces:
\begin{equation}
    R^t = \bigl[\,
        I - \hat{I}^t ;
        \phi(I) - \phi(\hat{I}^t)
    \,\bigr]
    \in \mathbb{R}^{(3+64) \times H \times W},
\end{equation}
where $\phi$ denotes the 64-channel relu1\_2 output of a frozen VGG-16 network and $[\cdot;\cdot]$ denotes concatenation. $R^t$ provides a 67-dimensional residual vector at each pixel location.
Each surviving Gaussian inherits the rendering error at the source-image pixel from which it was instantiated. Since the $K$ Gaussians along a camera ray originate from the same pixel, they receive identical residual features. Gathering this residual for each of the $m$ surviving Gaussians forms $E^t \in \mathbb{R}^{m \times 67}$.
We strictly restrict the recurrent feedback to the source view, as this is the only image available during inference.

\subsection{Training Objective}
\label{sec:loss}
We train the recurrent refinement module by supervising the rendered image at each iteration against both the source view and a held-out target view randomly sampled from the same scene.

\vspace{0.5em}
\noindent\textbf{Per-step Loss.}
For a given iteration $t$ and viewpoint $v \in \{\text{src}, \text{tgt}\}$, let $\hat{I}^t_v$ denote the image rasterized from the current Gaussian set $G^t$, and $I_v$ the corresponding ground-truth image. We compute the per-step loss as a weighted combination of pixel-wise and perceptual errors:
\begin{equation}
    \mathcal{L}_v(t) \;=\; \lambda_1\,\mathcal{L}_\text{MSE}(\hat{I}^t_v, I_v) + \lambda_2\,\mathcal{L}_\text{LPIPS}(\hat{I}^t_v, I_v).
\end{equation}
We use $\mathcal{L}_\text{src}(t)$ and $\mathcal{L}_\text{tgt}(t)$ to denote this loss evaluated at the source and target view, respectively.

\vspace{0.5em}
\noindent\textbf{Step-wise weighted loss.}
Since the network is unrolled for $T$ iterations, we accumulate the per-step losses with a geometric decay $\gamma \in (0, 1]$. This strategy ensures intermediate iterations remain well-supervised while strongly emphasizing the final output:
\begin{equation}
    \mathcal{L} \;=\; \sum_{t=1}^{T}
        \gamma^{\,T - t}
        \Bigl(
            \mathcal{L}_\text{src}(t)
          + \lambda_\text{tgt}\,\mathcal{L}_\text{tgt}(t)
        \Bigr).
\end{equation}

\vspace{0.5em}
\noindent\textbf{Hyperparameters.}
For the pruning mechanism, we set the importance score weights to $\lambda_\text{edge} = 2.0$ and $\lambda_\text{op} = 1.0$. For the refinement objective, we set the error weights to $\lambda_1 = 1.0$, $\lambda_2 = 0.5$, $\lambda_\text{tgt} = 0.5$, and the decay factor to $\gamma = 0.9$. To stabilize the early phase of training, $\lambda_\text{tgt}$ follows a three-phase warm-up schedule. Further training details are provided in Sec.~\ref{sec:implementation}.

\section{Experiments}
\label{sec:experiments}
\subsection{Experimental Settings}
\label{subsec:exp_settings}

\vspace{0.5em}
\noindent\textbf{Datasets.}
To evaluate our proposed method, we train our recurrent refinement module and conduct in-domain evaluation on the RealEstate10K dataset \cite{zhouStereoMagnificationLearning2018c}. We strictly follow the standard training and testing split used for Flash3D \cite{szymanowiczFlash3dFeedforwardGeneralisable2025}.
To assess the cross-domain generalization of our learned refinement module, we evaluate the models' zero-shot performance on the KITTI \cite{Geiger2012CVPR, Geiger2013IJRR} and DL3DV \cite{ling2024dl3dv} datasets. Details of the testing splits and view selection protocols are provided in the supplementary material. 

\vspace{0.5em}
\noindent\textbf{Evaluation Metrics.}
We adopt standard novel view synthesis metrics to assess rendering quality: PSNR, SSIM, and LPIPS \cite{zhangUnreasonableEffectivenessDeep2018e}.
We also report the number of Gaussians representing the scene, and primitive storage amount. 

\vspace{0.5em}
\noindent\textbf{Base Architectures.}
To demonstrate the general applicability of our method, 
we adopt three state-of-the-art single-view feed-forward 3DGS methods as our base architectures: Flash3D \cite{szymanowiczFlash3dFeedforwardGeneralisable2025}, Niagara \cite{wuNiagaraNormalIntegratedGeometric2025a}, and SHARP \cite{meschederSharpMonocularView2026}. Additionally, we use Flash3D as our primary architecture for detailed analysis and cross-domain evaluation, given its established evaluation protocols.

\vspace{0.5em}
\noindent\textbf{Evaluated Configurations (Baselines).} 
For each base architecture, we evaluate and compare the following three configurations to validate the effectiveness of our approach:
\begin{itemize}
    \item \textbf{Original:} The unmodified base model using 100\% of the predicted Gaussians. This serves as the upper-bound reference for rendering quality.
    \item \textbf{\ours{} (RM):} Random Mask (RM) is a controlled baseline in which 50\% of the Gaussians are pruned randomly. To strictly compare IM and RM, RM randomly samples the mask while keeping the layer-wise keep ratio identical to IM's. We then apply our recurrent refinement module.
    \item \textbf{\ours{} (IM):} Our proposed method retains 50\% of the Gaussians based on the Importance Mask (IM) pruning strategy, followed by our recurrent refinement.
\end{itemize}
We do not compare our methods with current compact feed-forward 3DGS methods (see examples of EcoSplat~\cite{parkEcoSplatEfficiencycontrollableFeedforward2026a} and F4Splat~\cite{kimF4SplatFeedForwardPredictive2026a} in Appendix~\ref{sec: compactff3dgs}).

\begin{table}[t]
\centering
\footnotesize
\caption{\textbf{Model-agnostic evaluation on RealEstate10K} \cite{zhouStereoMagnificationLearning2018c}. Comparison of original + \ours{} ($T=4$) using IM and RM at a keep ratio of $r=0.5$ across three single-view feed-forward 3DGS backbones (Flash3D \cite{szymanowiczFlash3dFeedforwardGeneralisable2025}, Niagara \cite{wuNiagaraNormalIntegratedGeometric2025a}, SHARP \cite{meschederSharpMonocularView2026}). The top row on each block represents the original base models without pruning or refinement. The \colorbox{red!40}{best} and \colorbox{red!20}{second-best} results per backbone are highlighted. \#G denotes the number of surviving Gaussians. Details of the primitive storage calculation are provided in the supplementary material (see Appendix~\ref{sec:storage}).}
\label{tab:indomain_results}
\resizebox{\columnwidth}{!}{%
\begin{tabular}{l ccc c c}
\toprule
Method & PSNR $\uparrow$ & SSIM $\uparrow$ & LPIPS $\downarrow$ & \#G (K) $\downarrow$ & Storage (MB) $\downarrow$  \\
\midrule
Flash3D~\cite{szymanowiczFlash3dFeedforwardGeneralisable2025}               & \cellcolor{red!40}25.10 & \cellcolor{red!40}0.834 & \cellcolor{red!40}0.155 & \cellcolor{red!20}286.72 & \cellcolor{red!20}26.38 \\
+ \ours{} (RM)        &  24.86 &  0.825 & \cellcolor{red!20}0.159 & \cellcolor{red!40}143.36 & \cellcolor{red!40}13.19 \\
+ \ours{} (IM)        & \cellcolor{red!20}25.04 & \cellcolor{red!20}0.826 & \cellcolor{red!40}0.155 & \cellcolor{red!40}143.36 & \cellcolor{red!40}13.19 \\
\midrule
Niagara~\cite{wuNiagaraNormalIntegratedGeometric2025a}               & \cellcolor{red!40}25.28 & \cellcolor{red!40}0.835 & \cellcolor{red!40}0.154 & \cellcolor{red!20}286.72 & \cellcolor{red!20}26.38 \\
+ \ours{} (RM)        &  24.57 &  0.813 &  0.168 & \cellcolor{red!40}143.36 & \cellcolor{red!40}13.19 \\
+ \ours{} (IM)        & \cellcolor{red!20}24.72 & \cellcolor{red!20}0.815 & \cellcolor{red!20}0.164 & \cellcolor{red!40}143.36 & \cellcolor{red!40}13.19 \\
\midrule
SHARP~\cite{meschederSharpMonocularView2026}                 & 24.89 & \cellcolor{red!20}0.828 & \cellcolor{red!40}0.142 & \cellcolor{red!20}1179.65 & \cellcolor{red!20}66.06 \\
+ \ours{} (RM)        & \cellcolor{red!20}25.17 & \cellcolor{red!40}0.829 & \cellcolor{red!40}0.142 & \cellcolor{red!40}589.82 & \cellcolor{red!40}33.03 \\
+ \ours{} (IM)        & \cellcolor{red!40}25.19 & \cellcolor{red!40}0.829 & \cellcolor{red!40}0.142 & \cellcolor{red!40}589.82 & \cellcolor{red!40}33.03 \\
\bottomrule
\end{tabular}%
}
\vspace{-8pt}
\end{table}

\vspace{0.5em}
\noindent\textbf{Implementation Details.} 
\label{sec:implementation}
For each frozen backbone, we train only the lightweight recurrent refinement module on its outputs. We optimize our refinement module using Adam \cite{kingma2015adam} with a learning rate of $1 \times 10^{-4}$ and a batch size of 2 for 30k iterations. We train the model on a single NVIDIA A100 GPU. During training, the number of recurrent unrolling steps $T$ is sampled uniformly from $\{1, \dots, 4\}$ to improve robustness to iterations, while we set $T = 4$ for evaluation. As mentioned in Sec.~\ref{sec:loss}, to stabilize the initial learning phase, the target view loss weight $\lambda_\text{tgt}$ follows a three-phase warm-up: $0$ for the first $10\text{k}$ iterations, $0.1$ until $20\text{k}$, and $0.5$ thereafter. We constrain the source-target frame distance to at most $30$ frames.

\subsection{Main Results}
\noindent\textbf{In-Domain Reconstruction}
\label{sec:indomain}
Table~\ref{tab:indomain_results} summarizes the quantitative results on the RealEstate10K \cite{zhouStereoMagnificationLearning2018c} dataset. We evaluate our compaction method across three distinct architectures: Flash3D \cite{szymanowiczFlash3dFeedforwardGeneralisable2025}, Niagara \cite{wuNiagaraNormalIntegratedGeometric2025a}, and SHARP \cite{meschederSharpMonocularView2026}, rendering at $256 \times 384$ resolution. Following the evaluation protocol of Flash3D, we select each target frame by uniformly sampling its temporal distances from the source view as $\mathcal{U}[-30, 30]$.

\vspace{0.5em}
Our proposed pipeline (+ \ours{} (IM)) successfully discards $50\%$ of the generated Gaussians while maintaining highly competitive visual fidelity compared to the base models. The results in Table~\ref{tab:indomain_results} show that the PSNR degradation is only $0.06$ dB for Flash3D and $0.56$ dB for Niagara. Using SHARP as a base model, the PSNR increases $0.30$ dB while reducing $50\%$ of the Gaussians. This demonstrates that our approach effectively reduces the number of Gaussians while maintaining robust rendering quality. 

To further examine the impact of the pruning strategy with recurrent refinement, we compare Original + \ours{} (IM) against Original + \ours{} (RM). Our compaction method using IM consistently outperforms the method using RM at the same compression rate, with the performance gap being most pronounced in Flash3D and Niagara. While the results are identical for SHARP, we attribute this to its exceptionally dense native point cloud ($>1$M primitives), where spatial redundancy remains extremely high even after a 50\% reduction at the $256 \times 384$ evaluation resolution. We verify this by measuring the pure mask effect at a lower keep ratio with no recurrent refinement (see Appendix~\ref{sec: sharp_T0}).

\begin{table}[t]
\centering
\caption{\textbf{Cross-dataset evaluation on KITTI \cite{Geiger2012CVPR,Geiger2013IJRR} and DL3DV \cite{ling2024dl3dv}.} Comparison of the baseline Flash3D~\cite{szymanowiczFlash3dFeedforwardGeneralisable2025} and Flash3D + \ours{} (RM and IM) at a keep ratio of $r=0.5$ ($T=4$). The \colorbox{red!40}{best} and \colorbox{red!20}{second-best} results per dataset are highlighted.}
\label{tab:flash3d_kitti_dl3dv}
\resizebox{\columnwidth}{!}{%
\setlength{\tabcolsep}{5pt}%
\begin{tabular}{l cccc cccc}
\toprule
\multirow{2}{*}{Method} & \multicolumn{4}{c}{KITTI\cite{Geiger2012CVPR, Geiger2013IJRR}} & \multicolumn{4}{c}{DL3DV\cite{ling2024dl3dv}} \\
\cmidrule(lr){2-5} \cmidrule(lr){6-9}
 & PSNR $\uparrow$ & SSIM $\uparrow$ & LPIPS $\downarrow$ & \#G (K) $\downarrow$ & PSNR $\uparrow$ & SSIM $\uparrow$ & LPIPS $\downarrow$ & \#G (K) $\downarrow$ \\
\midrule
Flash3D~\cite{szymanowiczFlash3dFeedforwardGeneralisable2025} & \cellcolor{red!40}21.88 & \cellcolor{red!40}0.827 & \cellcolor{red!40}0.130 & \cellcolor{red!20}98.30 & \cellcolor{red!40}20.47 & \cellcolor{red!40}0.616 & \cellcolor{red!40}0.253 & \cellcolor{red!20}245.76\\
+ \ours{} (RM) & 20.75 & 0.770 & 0.185 & \cellcolor{red!40}49.15 & 20.28 & 0.603 & 0.277 & \cellcolor{red!40}122.88\\
+ \ours{} (IM) & \cellcolor{red!20}21.08 & \cellcolor{red!20}0.778 & \cellcolor{red!20}0.176 & \cellcolor{red!40}49.15 & \cellcolor{red!20}20.35 & \cellcolor{red!20}0.605 & \cellcolor{red!20}0.267 & \cellcolor{red!40}122.88\\
\bottomrule
\end{tabular}%
}
\vspace{-8pt}
\end{table}

\vspace{0.5em}
\noindent\textbf{Cross-Domain Generalization}
To demonstrate that our compaction method retains the original backbone's zero-shot generalization, we evaluate models trained on RealEstate10K without dataset-specific fine-tuning on two datasets outside the training distribution: KITTI~\cite{Geiger2012CVPR,Geiger2013IJRR} (outdoor driving scenes, evaluated at $128\times384$ following the protocol of Flash3D), and DL3DV~\cite{ling2024dl3dv} (diverse real-world scenes, evaluated at $256\times480$). We focus on Flash3D for this evaluation because it provides an established cross-domain protocol that we follow directly on KITTI.

As reported in Table~\ref{tab:flash3d_kitti_dl3dv}, our framework retains competitive rendering quality while reducing the number of Gaussian primitives by 50\% on both datasets. Specifically, Flash3D + \ours{} (IM) achieves PSNR scores of 21.08 dB on KITTI and 20.35 dB on DL3DV, remaining within 0.80 dB and 0.12 dB, respectively, of the unpruned Flash3D baseline. Moreover, Flash3D + \ours{} (IM) consistently outperforms the corresponding Flash3D + \ours{} (RM) baseline across all three image-quality metrics, improving PSNR by 0.33 dB on KITTI and 0.07 dB on DL3DV while also achieving higher SSIM and lower LPIPS. These results indicate that importance-based Gaussian selection generalizes more effectively than random selection beyond the training distribution.

\begin{table}[t]
\centering
\footnotesize
\caption{\textbf{Comparison of Post-Hoc Pruning.} Comparison of our IM against RM and other post-hoc pruning methods ($r=0.5$, $T=0$, Dataset: RealEstate10K~\cite{zhouStereoMagnificationLearning2018c}). The \colorbox{red!40}{best} and \colorbox{red!20}{second-best} results are highlighted.}
\label{tab:pruning}
\resizebox{\columnwidth}{!}{%
\begin{tabular}{l cccc}
\toprule
Method & PSNR $\uparrow$ & SSIM $\uparrow$ & LPIPS $\downarrow$ & \#G (K) $\downarrow$ \\
\midrule
Flash3D~\cite{szymanowiczFlash3dFeedforwardGeneralisable2025}                 & \cellcolor{red!40}25.10 & \cellcolor{red!40}0.834 & \cellcolor{red!40}0.155 & \cellcolor{red!20}286.72 \\
Flash3D~\cite{szymanowiczFlash3dFeedforwardGeneralisable2025} + LightGaussian~\cite{fan2024lightgaussian}   & 24.08 & 0.811 & 0.226 & \cellcolor{red!40}143.36 \\
Flash3D~\cite{szymanowiczFlash3dFeedforwardGeneralisable2025} + PUP 3D-GS~\cite{hanson2025pup}        & 24.15 & 0.803 & 0.244 & \cellcolor{red!40}143.36 \\
Flash3D~\cite{szymanowiczFlash3dFeedforwardGeneralisable2025} + RM              & 23.92 & 0.798 & 0.260 & \cellcolor{red!40}143.36 \\
Flash3D~\cite{szymanowiczFlash3dFeedforwardGeneralisable2025} + IM             & \cellcolor{red!20}24.39 & \cellcolor{red!20}0.812 & \cellcolor{red!20}0.218 & \cellcolor{red!40}143.36 \\
\bottomrule
\end{tabular}
}
\vspace{-8pt}
\end{table}

\begin{figure}[t] 
  \centering
  \includegraphics[width=\columnwidth]{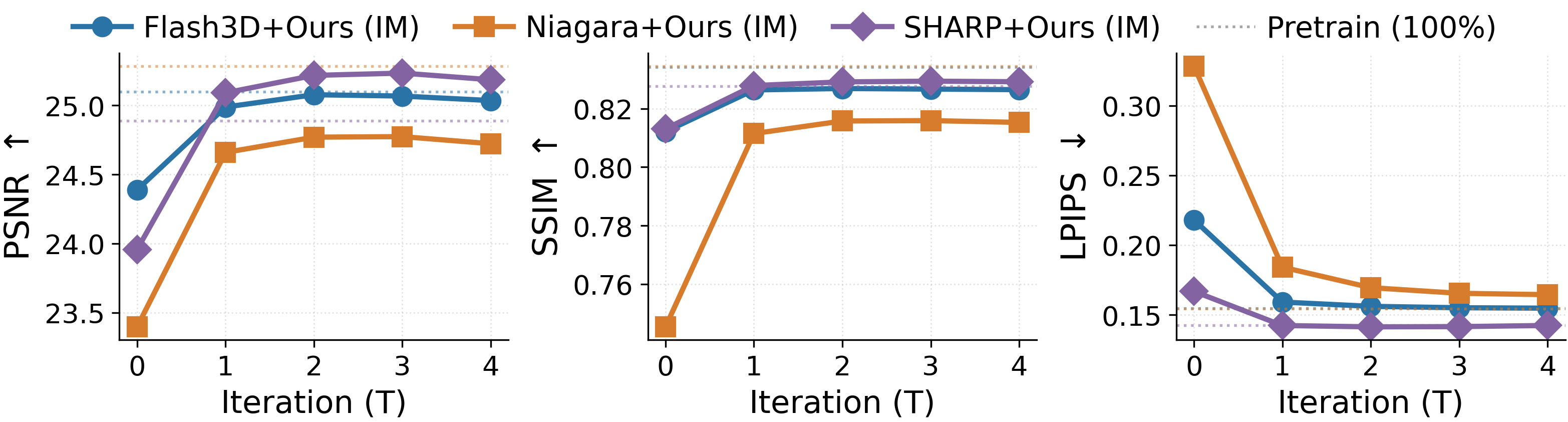}
  \caption{\textbf{Progressive refinement of Original + \ours{} (IM).} Evolution of PSNR, SSIM, and LPIPS from iterations $T=0$ to $T=4$ across three different backbones. The recurrent module consistently restores rendering fidelity ($r=0.5$, Dataset: RealEstate10K~\cite{zhouStereoMagnificationLearning2018c}). Dotted lines indicate the original models.}
  \label{fig:ProgressiveRefinement}
  \vspace{-8pt}
\end{figure}

\begin{figure*}[t]
\centering
\includegraphics[width=\textwidth]{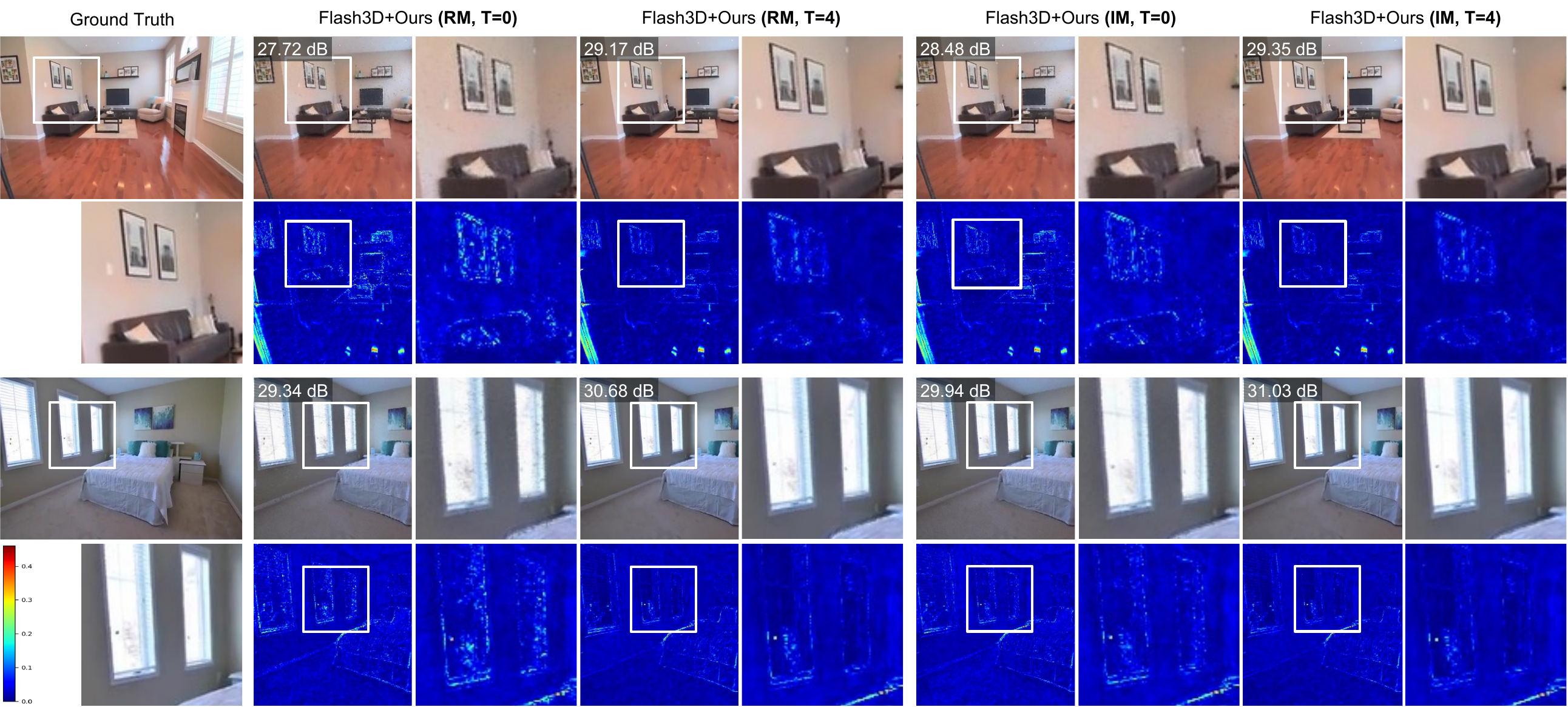}
\caption{\textbf{Qualitative progression of recurrent refinement.} Visualizing the rendered target views and corresponding error maps across unrolling iterations ($r=0.5$, Dataset: RealEstate10K~\cite{zhouStereoMagnificationLearning2018c}). From left to right: Ground Truth, RM ($T=0,4$), and IM ($T=0,4$).}
\label{fig:compare}
\vspace{-8pt}
\end{figure*}

\begin{figure}[t]
  \centering
  \includegraphics[width=\columnwidth]{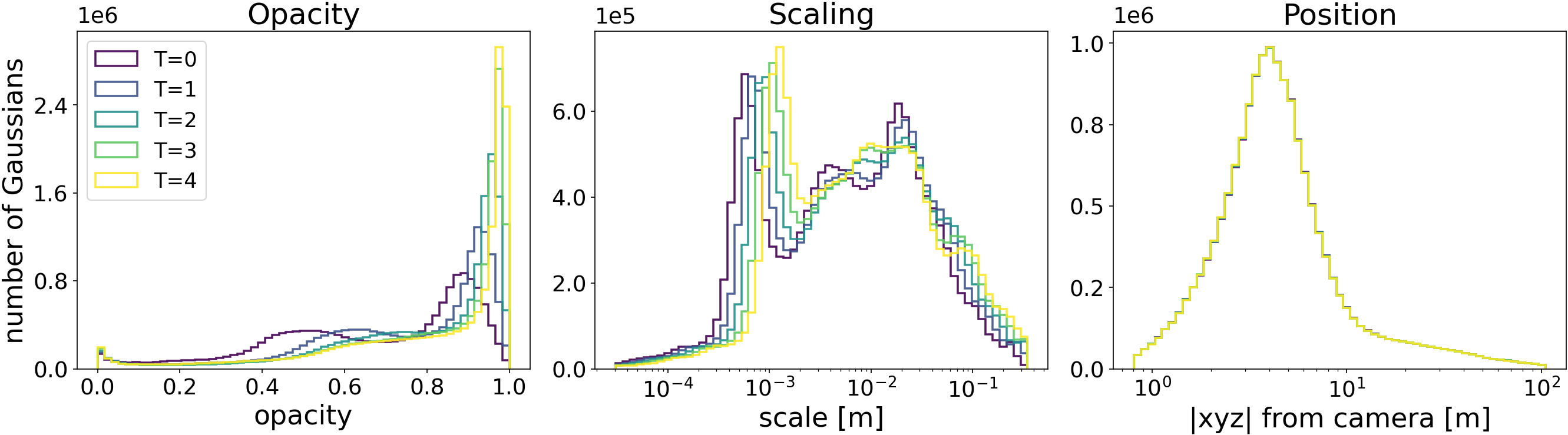}
  \caption{\textbf{Distribution of Gaussian parameters across recurrent iterations.} The graphs show the distribution of opacity, scaling, and position values of the Gaussian parameter at every iteration $T$. $T=0$ is the initial prediction after IM pruning, and iterations $T=1,2,3,4$ are the successive recurrent updates (Backbone: Flash3D~\cite{szymanowiczFlash3dFeedforwardGeneralisable2025}, $r=0.5$, Dataset: RealEstate10K~\cite{zhouStereoMagnificationLearning2018c}).}
  \label{fig:distribution}
  \vspace{-8pt}
\end{figure}

\subsection{Component Analysis}
\label{sec: comp-ana}
To provide deeper insights into the behavior of our pruning and refinement pipeline, we conduct extensive analysis using Flash3D as the representative base model.

\vspace{0.5em}
\noindent\textbf{Comparison of Post-Hoc Pruning.}
To validate our importance-based selection, we compare it against existing post-hoc pruning criteria, such as LightGaussian \cite{fan2024lightgaussian} and PUP 3D-GS \cite{hanson2025pup}, without recurrent refinement. We replace our importance score with the scores from LightGaussian and PUP 3D-GS, while keeping the rest of the pruning stage identical. We verified that stochastic sampling introduces negligible variance at the dataset level across $3205$ scenes, and that repeated evaluations with different random seeds differ by less than $0.005$ dB in PSNR. As shown in Table~\ref{tab:pruning}, our Importance Mask (IM) achieves better rendering quality than the other post-hoc pruning methods. Notably, all informed criteria clearly outperform random selection, and IM improves further on both, leading LightGaussian by $0.31$ dB and PUP 3D-GS by $0.24$ dB in PSNR; our simple edge–opacity score, computed directly from the input image and predicted opacities, outperforms criteria originally designed for optimized multi-view 3DGS. IM remains the best when the recurrent refinement module is trained on top of each criterion (see Appendix~\ref{sec: post-hoc}).

\vspace{0.5em}
\noindent\textbf{Progressive Recovery and Qualitative Validation.}
To explicitly quantify the repair capability of our recurrent refinement module, \figurename ~\ref{fig:ProgressiveRefinement} illustrates the progression of all metrics across recurrent iterations ($T=0$ to $4$) for all three backbones. Immediately after IM pruning ($T=0$), the rendering quality inherently drops. However, the recurrent network rapidly bridges this quality gap, acting as a robust, backbone-agnostic restorer.
Using Flash3D as a representative backbone, we visually validate this recovery process in \figurename ~\ref{fig:compare}. While RM uniformly discards primitives and induces severe structural degradation at $T=0$, IM deliberately preserves perceptually critical boundaries. As the refinement module processes the surviving primitives ($T=4$), it successfully interpolates and fills the spatial gaps, recovering the discarded image quality for high-fidelity novel view synthesis.

\vspace{0.5em}
\noindent\textbf{Analysis of Gaussian parameters across refinement.}
To analyze the behavior of our recurrent refinement module, we collected the distribution of Gaussian parameters across recurrent iterations (\figurename ~\ref{fig:distribution}). For each test scene, we randomly sample $5k$ retained Gaussians and aggregate them over the entire RealEstate10K test set. Each graph shows that, during recurrent iteration updates, the distribution of opacity and scaling values shifts to the right, while position values show minimal movement. This result shows that our recurrent refinement module tends to inflate the Gaussian parameters, making them more opaque and larger, to account for the IM pruning. Further results in Appendix~\ref{sec: para_analysis}.

\begin{figure}[t]
  \centering
  \includegraphics[width=\columnwidth]{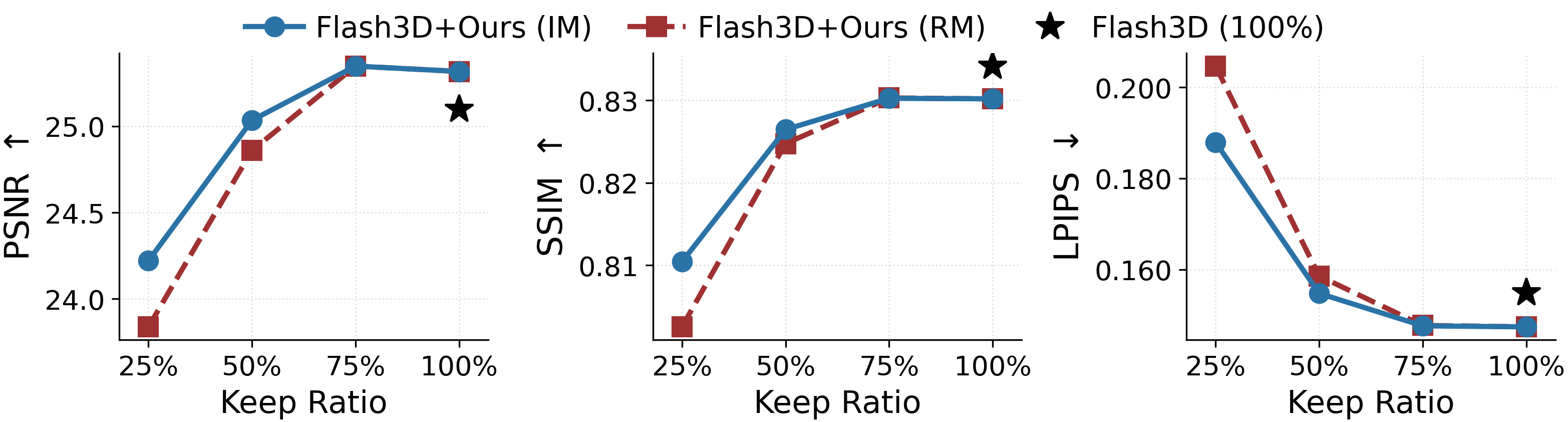}
  \caption{\textbf{Rate-distortion evaluation.} Comparison of Flash3D + \ours{} (IM) and Flash3D + \ours{} (RM) across varying keep ratios ($T=4$, Dataset: RealEstate10K~\cite{zhouStereoMagnificationLearning2018c}). The black star represents the original backbone. Note that IM and RM are identical at $r=1.0$, where no pruning is applied and only the refinement operates.}
  \label{fig:rd}
\vspace{-8pt}
\end{figure}

\begin{figure*}[t] 
  \centering
  \includegraphics[width=\textwidth]{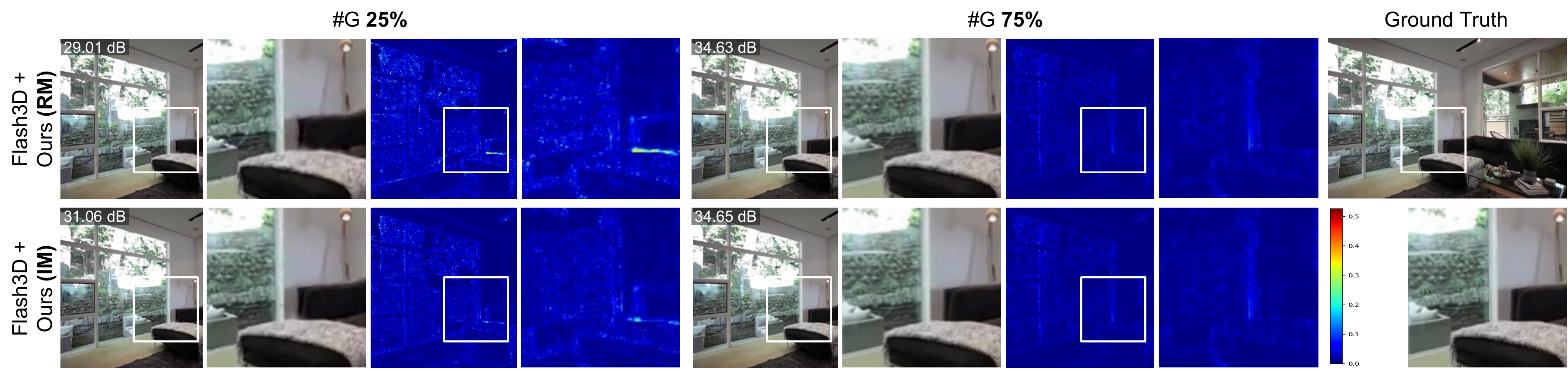}
  \caption{\textbf{Qualitative evaluation across different keep ratios} ($r=0.25,0.75$). Visual comparison of Flash3D + \ours{} (RM, $T=4$) and Flash3D + \ours{} (IM, $T=4$) on the RealEstate10K~\cite{zhouStereoMagnificationLearning2018c} dataset.}
  \label{fig:qualitative}  
  \vspace{-8pt}
\end{figure*}

\vspace{0.5em}
\noindent\textbf{Rate-Distortion Performance.} 
We trained and evaluated our method across keep ratios $r \in \{0.25, 0.5, 0.75, 1.0\}$. \figurename~\ref{fig:rd} shows the corresponding rate-distortion curves, where Flash3D + \ours{} (IM) consistently outperforms Flash3D + \ours{} (RM) under pruning. At $r=1.0$, no pruning is applied, and IM and RM are therefore identical. The performance gap between RM and IM is largest at lower keep ratios and becomes marginal at $r=0.75$, demonstrating the effectiveness and scalability of the proposed masking strategy.
Moreover, \figurename~\ref{fig:rd} shows that the refined outputs at $r \in \{0.75, 1.0\}$ outperform the original model in terms of both PSNR and LPIPS. These results indicate that recurrent refinement improves rendering quality, particularly under aggressive pruning, by substantially reducing the number of Gaussians while mitigating fidelity loss. The qualitative comparisons in \figurename~\ref{fig:qualitative} further show that IM maintains better visual quality than RM as the retained Gaussian budget decreases from $75\%$ to $25\%$, consistent with the smaller PSNR degradation of IM observed in \figurename~\ref{fig:rd}. Overall, our method using IM effectively retains rendering quality at low keep ratios and achieves a favorable trade-off between compaction and high-fidelity synthesis.

\subsection{Robustness to Unseen Keep Ratios.} 
In practical deployment, the desired keep ratio may vary depending on the target application's storage budget or quality requirements. To avoid re-training for every possible ratio, we investigate whether a single model can flexibly support multiple keep ratios at inference time. Specifically, we evaluate the zero-shot generalization of a recurrent refinement module trained at a fixed 50\% keep ratio when tested at 25\% and 75\%. Furthermore, we introduce \textit{Keep Ratio Mix} (krMix), a training strategy in which the keep ratio is randomly sampled from $r \in \{0.25, 0.5, 0.75\}$ at each training step. 
As shown in Table~\ref{tab:kra}, the model trained with a fixed 50\% keep ratio exhibits asymmetric generalization. It transfers reasonably well to a keep ratio of $0.25$, but suffers a substantial PSNR degradation at $0.75$. The three models produce identical outputs at $T=0$, as the mask does not depend on the refinement module; the gap emerges only as refinement iterates, showing that errors on unseen primitive configurations compound across steps. In contrast, krMix avoids this failure, staying within $0.17$ dB of ratio-specific training across all ratios while achieving the best PSNR and SSIM at the test keep ratio of $0.25$. These results demonstrate that exposure to each keep ratio during training is crucial, and that a single krMix-trained model can flexibly support all evaluated keep ratios while matching or exceeding ratio-specific training.

\begin{table}[t]
\centering
\footnotesize
\caption{\textbf{Robustness to unseen keep ratios.} Evaluation of different training strategies for Flash3D + \ours{} (IM, $T=4$) (Dataset: RealEstate10K\cite{zhouStereoMagnificationLearning2018c}). We compare \textit{krMix} against models trained specifically for each test ratio (\textit{Ratio-Specific}) and a baseline trained solely at $r=0.5$ (\textit{Fixed 50\%}). The \colorbox{red!40}{best} and \colorbox{red!20}{second-best} results per ratio are highlighted.}
\label{tab:kra}
\resizebox{\columnwidth}{!}{%
\begin{tabular}{l ccc ccc ccc}
\toprule

& \multicolumn{3}{c}{Test keep ratio: $0.25$}
& \multicolumn{3}{c}{Test keep ratio: $0.5$}
& \multicolumn{3}{c}{Test keep ratio: $0.75$} \\

\cmidrule(lr){2-4}
\cmidrule(lr){5-7}
\cmidrule(lr){8-10}

Training Strategy
& PSNR $\uparrow$
& SSIM $\uparrow$
& LPIPS $\downarrow$
& PSNR $\uparrow$
& SSIM $\uparrow$
& LPIPS $\downarrow$
& PSNR $\uparrow$
& SSIM $\uparrow$
& LPIPS $\downarrow$ \\

\midrule
Ratio-Specific & 24.22 & \cellcolor{red!20}0.810 & \cellcolor{red!40}0.188 & \cellcolor{red!40}25.04 & \cellcolor{red!40}0.826 & \cellcolor{red!40}0.155 & \cellcolor{red!40}25.35 & \cellcolor{red!40}0.830 & \cellcolor{red!40}0.148 \\
Fixed 50\% & \cellcolor{red!20}24.29 & 0.804 & 0.220 & \cellcolor{red!40}25.04 & \cellcolor{red!40}0.826 & \cellcolor{red!40}0.155 & 22.35 & 0.808 & 0.176 \\
krMix & \cellcolor{red!40}24.35 & \cellcolor{red!40}0.811 & \cellcolor{red!20}0.191 & \cellcolor{red!20}24.93 & \cellcolor{red!20}0.825 & \cellcolor{red!20}0.157 & \cellcolor{red!20}25.18 & \cellcolor{red!20}0.830 & \cellcolor{red!20}0.149 \\

\bottomrule
\end{tabular}%
}
\vspace{-8pt}
\end{table}

\subsection{Ablation Studies}
\label{subsec:ablation}
We ablate the key components of our pipeline to validate our design choices. All ablation experiments are conducted using Flash3D on the RealEstate10K dataset.

\vspace{0.5em}
\noindent\textbf{Importance Metrics.} 
First, we perform an ablation on the importance metric. Our Importance Mask (IM) score is calculated with a combination of edge and opacity values. We replace this score with variants using only edge, only opacity, or SparseSplat's \cite{zhangSparseSplatApplicableFeedForward2026a} entropy. As shown in the top section of Table~\ref{tab:ablation}, the results show that alternative metrics relying solely on Edge, Opacity, or Entropy perform worse than our Importance Mask. Although opacity alone achieves comparable results, incorporating the edge score helps retain more important Gaussians, and this advantage becomes more pronounced at a keep ratio of $0.25$.

\vspace{0.5em}
\noindent\textbf{Recurrent Network Inputs.} 
Next, we investigate the impact of the input features provided to the recurrent refinement module. As reported in the middle section of Table~\ref{tab:ablation}, removing the hidden-state or the source view error causes a slight performance drop at both keep ratios $ 0.25$ and $ 0.5$. This confirms that these features provide useful context and error-correcting cues that support the recurrent refinement module to accurately refine the Gaussian parameters.

\begin{table}[t]
\centering
\caption{\textbf{Ablation Studies.} Evaluation on the Flash3D~\cite{szymanowiczFlash3dFeedforwardGeneralisable2025} backbone ($r=0.25, 0.5$, $T=4$, Dataset: RealEstate10K~\cite{zhouStereoMagnificationLearning2018c}). Top: Importance metric variants. Middle: Recurrent network input signals. Bottom: Our full proposed configuration. The \colorbox{red!40}{best} and \colorbox{red!20}{second-best} results are highlighted.}
\label{tab:ablation}
\resizebox{\columnwidth}{!}{%
\begin{tabular}{l ccc ccc }
\toprule
\multirow{2}{*}{Method} & \multicolumn{3}{c}{Keep ratio: $0.25$} & \multicolumn{3}{c}{Keep ratio: $0.5$} \\
\cmidrule(lr){2-4} \cmidrule(lr){5-7}
 & PSNR $\uparrow$ & SSIM $\uparrow$ & LPIPS $\downarrow$ & PSNR $\uparrow$ & SSIM $\uparrow$ & LPIPS $\downarrow$ \\
\midrule
\textbf{Importance Metric} &  &  &  &  &  &  \\
Edge              & 23.07 & 0.778 & 0.225 & 24.46 & 0.820 & 0.164 \\
Opacity           & 24.07 & \cellcolor{red!20}0.809 & \cellcolor{red!20}0.193 & 24.98 & \cellcolor{red!20}0.826 & \cellcolor{red!20}0.156 \\
Entropy           & 23.42 & 0.797 & 0.213 & 24.48 & 0.821 & 0.164 \\
\midrule
\textbf{Recurrent Network Inputs} &  &  &  &  &  &  \\
w/o hidden-state       & \cellcolor{red!20}24.13 & 0.808 & 0.201 & \cellcolor{red!20}24.99 & 0.825 & 0.158 \\
w/o src view error     & 23.32 & 0.790 & 0.232 & 24.52 & 0.821 & 0.165 \\
\midrule
Flash3D + \ours{} (IM)             & \cellcolor{red!40}24.22 & \cellcolor{red!40}0.810 & \cellcolor{red!40}0.188 & \cellcolor{red!40}25.04 & \cellcolor{red!40}0.826 & \cellcolor{red!40}0.155 \\
\bottomrule
\end{tabular}
}
\vspace{-8pt}
\end{table}

\section{Conclusion}
In this paper, we addressed the spatial and memory redundancies caused by the $K$-Gaussians-per-pixel paradigm in single-view feed-forward 3DGS. We proposed a post-hoc approach that integrates importance-based spatial pruning with a lightweight recurrent refinement module. Our compaction method can discard an arbitrary number of generated primitives while maintaining rendering quality. Furthermore, by introducing the krMix training strategy, a single recurrent model generalizes to unseen keep ratios, enabling flexible, dynamic adjustment of the keep ratio during inference without retraining. Ultimately, this post-hoc pipeline provides a highly scalable solution that 
facilitates storage-efficient 3DGS representations.

{
    \small
    \bibliographystyle{ieeenat_fullname}
    \bibliography{main}
}
\clearpage
\setcounter{page}{1}
\maketitlesupplementary

\section{Implementation Details}
\subsection{Details of Evaluation}
We evaluate the RealEstate10K-trained Flash3D backbone and its pruned variants directly on RealEstate10K, KITTI, and DL3DV. Following Flash3D's evaluation protocol, 5\% of the image border is cropped during evaluation. For KITTI, we follow the original Flash3D cross-domain evaluation protocol. Specifically, we use the Tulsiani split, which all test samples are indexed by the left view, and evaluation is performed on the corresponding stereo target view. For DL3DV, we adopt the view-indexing convention from DepthSplat~\cite{xuDepthsplatConnectingGaussian2025}. The original DepthSplat index contains two source views of indices $0$ and $9$, with target views of indices $1,3,5,7$. Since our evaluation uses the single-view setting, we only use the first source view with index $0$, while retaining the same four target views. 

\subsection{Model Configurations and Storage Calculation}
\label{sec:storage}
\paragraph{Note on model parameter count.}
The Gaussian parameter dimensionality $C$ depends on the base model and denotes the number of stored parameters for each primitive (Table~\ref{tab:gauss_param}).
For Flash3D, the model outputs $24$ channels per pixel. This comprises opacity (1), depth $d$ (1), pixel offset $\Delta \in \mathbb{R}^3$, scale (3), rotation (4), and SH coefficients (12, split into 3 for DC and 9 for the rest). The depth and offset are immediately consumed along with the 2D pixel coordinate $u$ to compute the 3D mean $\mu = K^{-1} u d + \Delta \in \mathbb{R}^3$ (the xyz position), where $K$ is the camera intrinsics. Therefore, the stored representation collapses to $C = 23$. Niagara follows the same convention. SHARP predicts 3D positions directly and uses RGB color without higher-order spherical harmonics, yielding $C = 14$.

\begin{table}[h]
\centering
\caption{Detail of the Gaussian parameters of each backbone.}
\label{tab:gauss_param}
\resizebox{\columnwidth}{!}{%
\setlength{\tabcolsep}{6pt}%
\begin{tabular}{lcl}
\toprule
Backbone & $C$ & Breakdown \\
\midrule
Flash3D & 23 & xyz(3) + opacity(1) + scale(3) + rot(4) + dc(3) + SH-rest(9) \\
Niagara & 23 & xyz(3) + opacity(1) + scale(3) + rot(4) + dc(3) + SH-rest(9) \\
SHARP   & 14 & xyz(3) + opacity(1) + scale(3) + rot(4) + RGB(3) \\
\bottomrule
\end{tabular}
}
\vspace{-8pt}
\end{table}

\paragraph{Storage calculation.}
Assuming float32 precision, primitive storage is calculated as 
$\mathrm{Storage} = n \times C \times 4$ bytes. The total number of Gaussians is $n = H \times W \times K$ for the baseline, where $H\times W$ denotes the dimensions of the output pixel grid and $K$ is the number of primitives per pixel. After pruning with a keep ratio $r$, the count becomes $n = \lfloor r \times H \times W \times K \rfloor$. Since the per-primitive parameter count $C$ is constant, the storage reduction scales directly with $r$.
Specifically, Flash3D and Niagara pad the input images by 32 pixels on all sides, resulting in an output pixel grid of $H\times W=320\times448$ with $K=2$. For SHARP, the Gaussian pixel grid is fixed at $H\times W=768\times768$ regardless of the input image resolution.   

\subsection{Hyperparameters for Recurrent Refinement}

The scaling vector $L$ in $G^{t+1} = G^t + L \odot \tanh(\Delta G^t)$ (Equation ~\ref{eq:update_g_z}) bounds the maximum per-step update for each Gaussian parameter. The values are calibrated in the activation space used by the backbone (e.g., opacity in logit space, scale in log space) so that no parameter exceeds its allowable range. Specifically, $L$ restricts opacity-logit updates to $0.5$, log-scale updates to $0.2$, and all remaining parameters to $0.1$ per iteration. The opacity bound of $0.5$ in logit space corresponds to a maximum additive change of approximately $0.12$ in the sigmoid-activated opacity, and the scale bound of $0.2$ in log space corresponds to a maximum multiplicative change of $\exp(0.2) \approx 1.22{\times}$ per step.

\begin{figure}[h]
  \centering
  \includegraphics[width=\columnwidth]{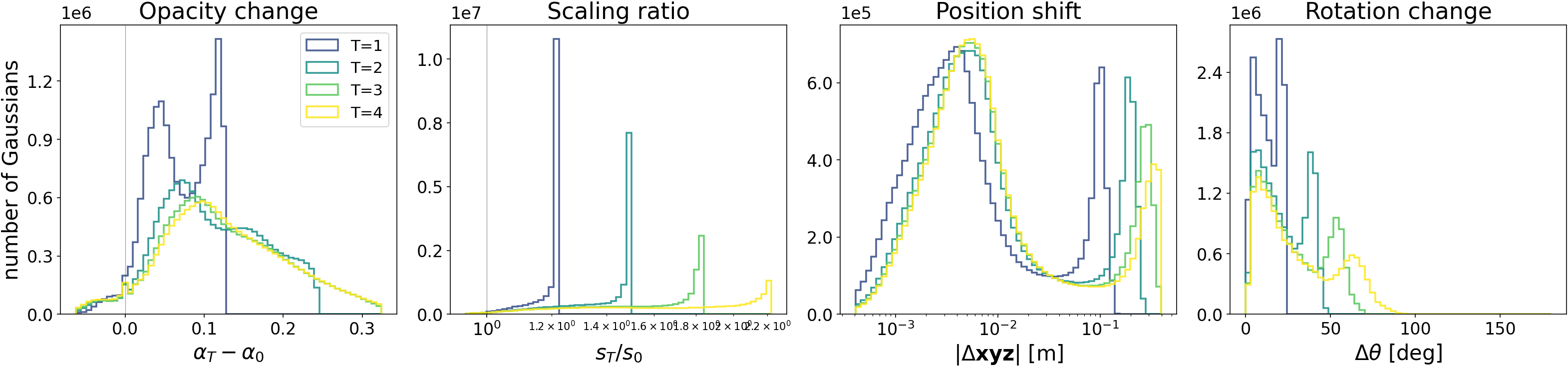}
  \caption{Distribution of Gaussian parameter updates across recurrent iterations.}
  \label{fig:delta_distribution}
\end{figure}

\begin{figure}[h]
  \centering
  \includegraphics[width=\columnwidth]{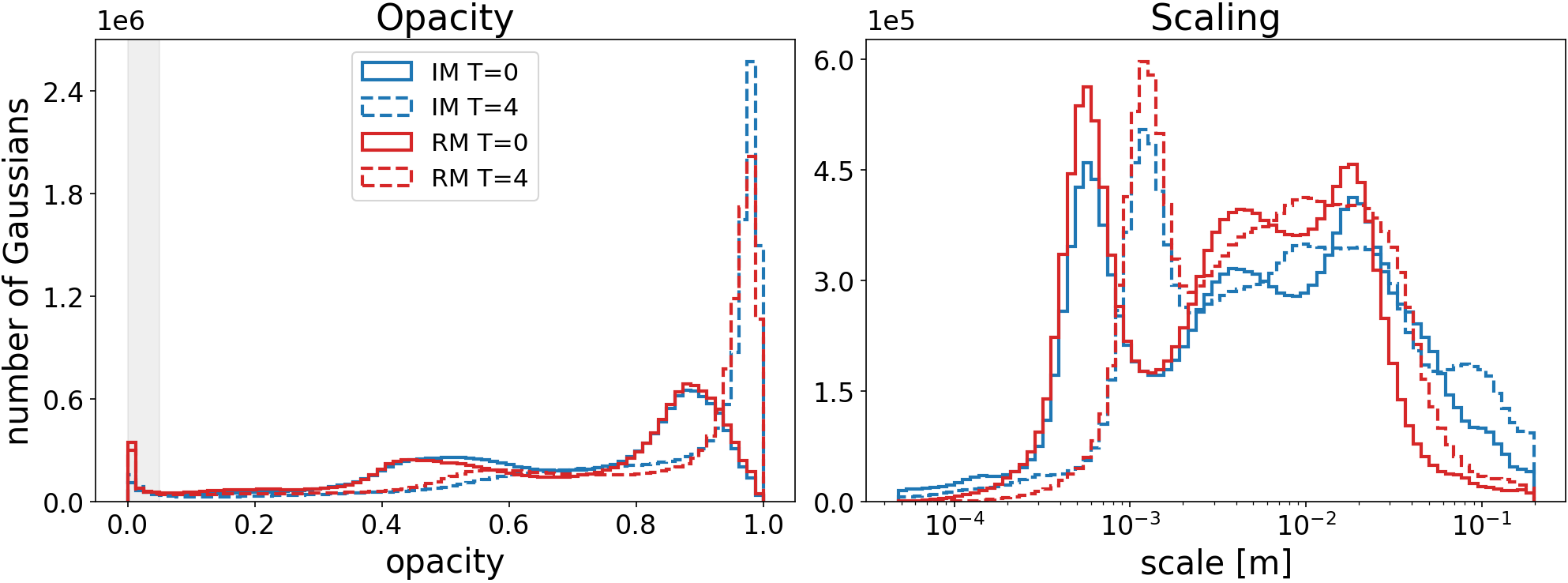}
  \caption{Comparison of Gaussian parameter distribution between IM and RM.}
  \label{fig:IM_RM_distribution}
\end{figure}

\section{Additional Architectural Analysis}
\subsection{Statistical Analysis of Parameter Updates}
\label{sec: para_analysis}
In \figurename ~\ref{fig:delta_distribution}, we show the distribution of Gaussian parameter delta values accumulated from the initial state across recurrent iterations, collected in the same way as \figurename ~\ref{fig:distribution}. The total change at iteration $T$ is calculated relative to the initial state ($T=0$). Opacity is shown as an additive difference ($\alpha_T-\alpha_0$), scaling as a multiplicative ratio ($s_T/s_0$ in log-space), position as the Euclidean displacement ($|\Delta \mathbf{xyz}|$), and rotation as the geodesic angle ($\Delta\theta$) between quaternions. 
The scaled tanh bounds each single step update preventing the recurrent network from changing a parameter too aggressively. The figure shows that the opacity, scaling, and rotation updates tend to accumulate near these bounds. On the other hand, the position shift remains small throughout the recurrent iteration, which is consistent with \figurename ~\ref{fig:distribution}, where the absolute values of the position show no difference.

Additionally, we compare the absolute opacity and scaling values for the IM and RM strategies (\figurename ~\ref{fig:IM_RM_distribution}). Both IM and RM exhibit similar distributional shifts as the number of recurrent update steps increases from $0$ to $4$. The primary difference is the percentage of low-opacity Gaussians. While the distribution collected by the RM strategy reflects the base model's output distribution, the IM strategy selects more Gaussians with higher opacity values.

\subsection{Visualization of Sampled Masks}
\label{sec: masks}
\begin{figure}[t] 
  \centering
  \includegraphics[width=\columnwidth]{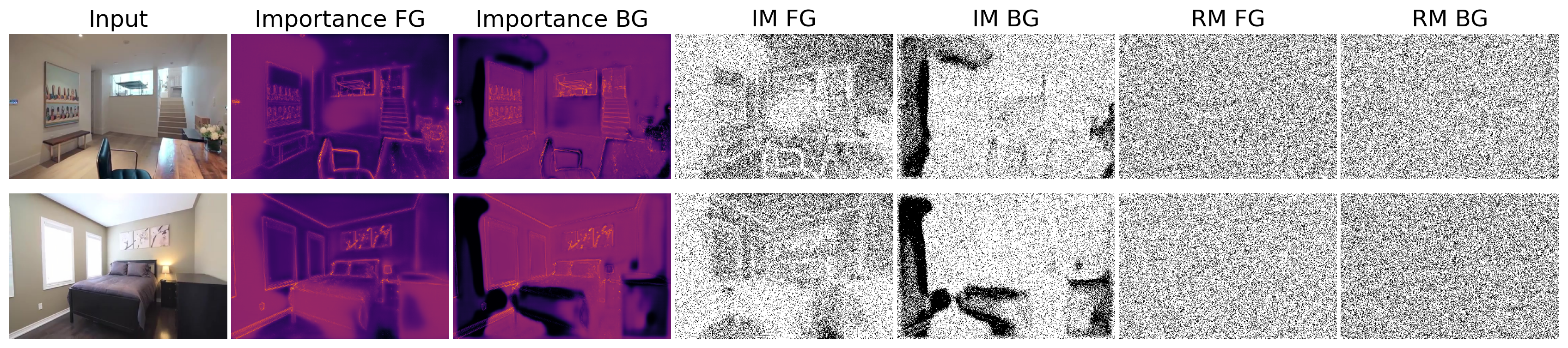}
  \caption{Visualization of the calculated importance map and sampled mask from the Flash3D backbone for both the front (FG) and back (BG) layers ($r=0.5$). Maps and masks are shown over the pixel grid of the source image.}
  \label{fig:mask_flash3d}
\end{figure}

\begin{figure}[t] 
  \centering
  \includegraphics[width=\columnwidth]{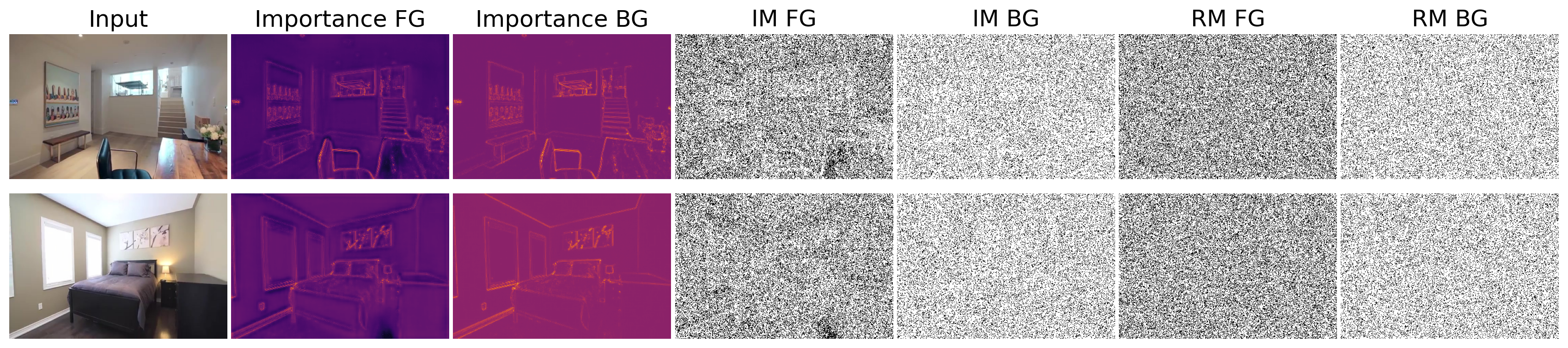}
  \caption{Visualization of the calculated importance map and sampled mask from the Niagara backbone for both the front (FG) and back (BG) layers ($r=0.5$). Maps and masks are shown over the pixel grid of the source image.}
  \label{fig:mask_niagara}
\end{figure}

\begin{figure}[t] 
  \centering
  \includegraphics[width=\columnwidth]{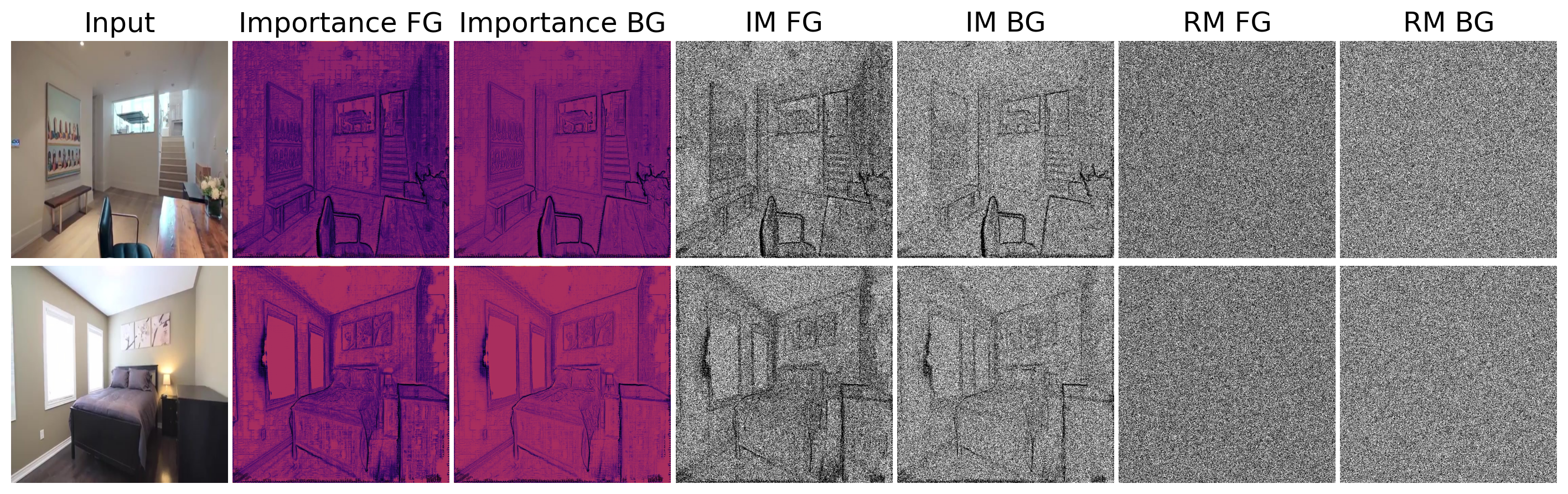}
  \caption{Visualization of the calculated importance map and sampled mask from the SHARP backbone for both the front (FG) and back (BG) layers ($r=0.5$). Maps and masks are shown over the pixel grid of the source image.}
  \label{fig:mask_sharp}
\end{figure}

In \figurename ~\ref{fig:mask_flash3d}, \ref{fig:mask_niagara}, and \ref{fig:mask_sharp}, we visualize the calculated importance map and sampled mask (IM, RM) with three different backbones. These visualizations demonstrate how our pruning framework adapts to the distinct architectural characteristics and geometric priors of each base model.

For Flash3D (\figurename ~\ref{fig:mask_flash3d}), the importance map shows higher scores for visible surfaces in the foreground. 
In the background, the importance map correctly assigns low scores to flat, occluded surfaces. Consequently, the sampled IM aggressively drops Gaussians in these redundant planar regions. In contrast, RM drops Gaussians randomly without preserving any geometric structure.

Regarding Niagara (\figurename ~\ref{fig:mask_niagara}), the background importance map appears uniformly flat. This occurs because Niagara's architecture tends to output background opacity values near $1.0$. While this model-specific characteristic leads to near-random sampling in the background, our IM slightly extracts geometric information from the foreground boundaries. This localized structural awareness allows IM to consistently outperform RM, despite the challenging background distribution.

Finally, SHARP (\figurename ~\ref{fig:mask_sharp}) shows a different architectural design, resizing inputs to a fixed square resolution. Consequently, its corresponding importance maps and masks are square. The visualization shows that SHARP tends to predict lower opacities near the edges of the image. Despite this architectural difference, the quantitative evaluation at $T=0$ (Table ~\ref{tab:sharpt0}) indicates that our IM maintains a performance advantage over RM, retaining Gaussians that contribute to rendering quality.

\subsection{Runtime Analysis}

We measure runtime on the RealEstate10K test set using an NVIDIA RTX
4070 Ti with a batch size of 1. We compare Flash3D with Flash3D + \ours{} (IM) using
$r=0.5$ and $T=4$. Each test scene contains one source view and three target views. The first 20 scenes are discarded as warm-up, leaving 3,185 samples for measurement. Data loading, host-to-device transfer, and
evaluation overhead are excluded.

Table~\ref{tab:runtime_breakdown} is organized into two parts. The upper
part reports the stage-wise runtime of the complete inference pipeline,
decomposed into five non-overlapping stages: Flash3D prediction, masking,
camera alignment, recurrent refinement, and output rendering. Masking
includes scoring, sampling, and mask application. The lower part reports
two throughput measurements obtained separately without stage-level
timing. Full-inference throughput measures the same complete inference pipeline,
from GPU-resident input to the standard outputs, while novel-view throughput measures repeated rendering with prepared Gaussian representations and target cameras.

\begin{table}[t]
\centering
\caption{
Runtime breakdown and throughput comparison of Flash3D~\cite{szymanowiczFlash3dFeedforwardGeneralisable2025} and Flash3D + \ours{} (IM) on the RealEstate10K~\cite{zhouStereoMagnificationLearning2018c} test set ($r=0.5$, $T=4$). Full-inference throughput measures the number of complete test scenes processed per second. Novel-view rendering throughput
measures target views rendered per second after reconstruction using prepared Gaussians and cameras. The two throughput metrics are measured 
separately.
}
\label{tab:runtime_breakdown}
\small
\resizebox{\columnwidth}{!}{%
\setlength{\tabcolsep}{4pt}
\begin{tabular}{lccc}
\toprule
Stage / metric
& Flash3D~\cite{szymanowiczFlash3dFeedforwardGeneralisable2025}
& Flash3D~\cite{szymanowiczFlash3dFeedforwardGeneralisable2025} + \ours{} (IM)
& Change \\
\midrule
\multicolumn{4}{l}{\textit{Stage runtime (ms/scene) $\downarrow$}} \\
Flash3D prediction & 124.45 & 123.00 & -- \\
Masking            & --     & 0.86   & -- \\
Camera alignment   & 76.64  & 74.14  & -- \\
RNN refinement     & --     & 76.07  & -- \\
Output rendering   & 5.64   & 4.94   & -- \\
\cmidrule(lr){1-4}
Total              & 207.13 & 279.53 & $+35.0\%$ \\
\midrule
\multicolumn{4}{l}{\textit{Throughput}} \\
Full-inference (scenes/s) $\uparrow$
& \textbf{4.85} & 3.36 & $-30.8\%$ \\
Novel-view (views/s) $\uparrow$
& 1,443 & \textbf{2,257} & $\mathbf{+56.4\%}$ \\
\bottomrule
\end{tabular}
}
\end{table}
Masking adds only 0.86 ms per scene, while four-step recurrent refinement adds 76.07 ms and dominates the additional runtime. Accordingly, Flash3D + \ours{} reduces full-inference throughput from 4.85 to 3.36 scenes/s. However, its compact representation increases novel-view rendering throughput from 1,443 to 2,257 views/s, corresponding to a 56.4\% improvement.

\begin{table}[t]
\centering
\caption{SHARP mask quality at T=0 (no recurrent refinement) under aggressive pruning.
PSNR, SSIM, and LPIPS on RealEstate10K~\cite{zhouStereoMagnificationLearning2018c} test set at keep ratios 25\%, 10\%, evaluated immediately after mask-based Gaussian pruning (T=0, no recurrent refinement). The best results are highlighted in \textbf{bold}.}
\label{tab:sharpt0}
\resizebox{\columnwidth}{!}{%
\setlength{\tabcolsep}{4pt}%
\begin{tabular}{l cccc}
\toprule
Method & PSNR $\uparrow$ & SSIM $\uparrow$ & LPIPS $\downarrow$ & \#G (K) $\downarrow$\\
\midrule
SHARP~\cite{meschederSharpMonocularView2026} + RM 25\% (T=0)            & 21.30 & 0.678 & 0.381 & 294.91\\
SHARP~\cite{meschederSharpMonocularView2026} + IM 25\% (T=0)            & \textbf{21.65} & \textbf{0.710} & \textbf{0.345} & 294.91 \\
\midrule
SHARP~\cite{meschederSharpMonocularView2026} + RM 10\% (T=0)            & 15.32 & 0.353 & 0.592 & 117.96\\
SHARP~\cite{meschederSharpMonocularView2026} + IM 10\% (T=0)            & \textbf{16.06} & \textbf{0.389} & \textbf{0.568} & 117.96\\
\bottomrule
\end{tabular}%
}
\end{table}

\section{Extended Results}
\subsection{SHARP mask quality at T=0}
\label{sec: sharp_T0}

To provide a deeper analysis of the SHARP backbone, we compared IM and RM pruning at low keep ratios, $r=0.10$ and $0.25$. Under this more aggressive pruning setting, IM consistently outperforms RM across all settings, as shown in Table~\ref{tab:sharpt0}. This supports our interpretation in the main paper (Sec.~\ref{sec:indomain}) that the smaller IM--RM gap at the 50\% reduction setting arises from SHARP's high spatial redundancy. 





\subsection{Results of Compact Feed-Forward 3DGS}
\label{sec: compactff3dgs}
As mentioned in related work (Sec.~\ref{sec: cff3dgs}), most work on feed-forwardly producing compact 3DGS representations is in the multi-view setting. EcoSplat~\cite{parkEcoSplatEfficiencycontrollableFeedforward2026a}, F4Splat~\cite{kimF4SplatFeedForwardPredictive2026a}, and SparseSplat~\cite{zhangSparseSplatApplicableFeedForward2026a} are the representative models that output a compact 3DGS from an arbitrary number of input images. SparseSplat does not release its code; therefore, we compared our method with EcoSplat and F4Splat in the same single-view setting to validate whether the current work can operate without multi-view information. Both methods are evaluated with the same settings as in Table~\ref{tab:indomain_results} ($r=0.5$, Dataset: RealEstate10K). EcoSplat outputs $0.5\times256\times384\times6$ number of Gaussians, where $6$ is the fixed number of anchor views into which EcoSplat's pretrained ZPressor backbone compresses the input view set. With a single input image, all six anchor slots are filled by that same view, so the output contains six replicated copies of the same Gaussian set. F4Splat requires a minimum of two input views, therefore we input two images from the same viewpoint to predict Gaussians. The model outputs $0.5\times384\times384\times2$ Gaussians. Its pre-trained model is trained only on square crops, so we pad the full frame to a square context, while novel views are still rendered and scored at $256\times384$ as in Table~\ref{tab:indomain_results}.

\begin{figure*}[t] 
  \centering
  \includegraphics[width=\linewidth]{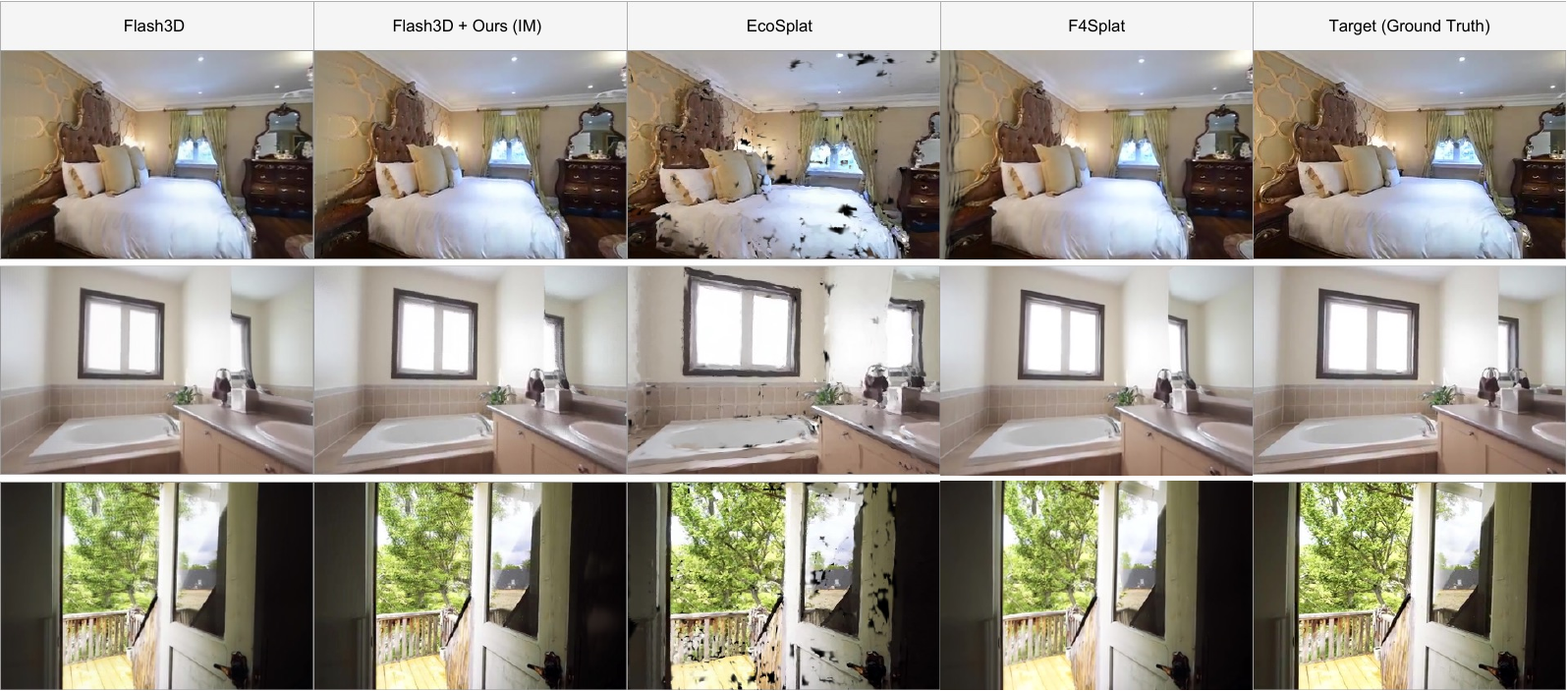}
  \caption{Comparison of Flash3D~\cite{szymanowiczFlash3dFeedforwardGeneralisable2025}, Flash3D + \ours{} (IM), EcoSplat~\cite{parkEcoSplatEfficiencycontrollableFeedforward2026a}, and F4Splat~\cite{kimF4SplatFeedForwardPredictive2026a}. Flash3D\cite{szymanowiczFlash3dFeedforwardGeneralisable2025} is shown without pruning as a reference; the other three use the same keep ratio $r=0.5$. The resulting Gaussian counts differ, since each architecture defines its budget over a different base grid.}
  \label{fig:eco_comp}
\end{figure*}

Looking at Table~\ref{tab:ecosplat}, there is a large gap in the rendering quality between EcoSplat and Flash3D + \ours{} (IM). Despite EcoSplat outputting more Gaussians than Flash3D + \ours{}, its rendering quality degrades markedly, with visible holes in Figure~\ref{fig:eco_comp}. For F4Splat, the gap to Flash3D + \ours{} (IM) is far smaller than for EcoSplat, but it still falls $2.2$ dB behind at a nearly identical Gaussian count. Taken together, multi-view compact methods can be run on a single input, but their effectiveness depends on the method. The absence of cross-view information shows a clear degradation. 

\begin{table}[t]
\centering
\caption{Comparison of Flash3D~\cite{szymanowiczFlash3dFeedforwardGeneralisable2025} and Flash3D + \ours{} (IM) with a compact feed-forward 3DGS method, EcoSplat~\cite{parkEcoSplatEfficiencycontrollableFeedforward2026a} ($r=0.5$, Dataset: RealEstate10K~\cite{zhouStereoMagnificationLearning2018c}). Flash3D + \ours{} (IM) is evaluated after $T=4$ recurrent refinements. The \colorbox{red!40}{best} and \colorbox{red!20}{second-best} results are highlighted.}
\label{tab:ecosplat}
\resizebox{\columnwidth}{!}{%
\setlength{\tabcolsep}{4pt}%
\begin{tabular}{l cccc}
\toprule
Method & PSNR $\uparrow$ & SSIM $\uparrow$ & LPIPS $\downarrow$  & \#G (K) $\downarrow$ \\
\midrule
Flash3D~\cite{szymanowiczFlash3dFeedforwardGeneralisable2025}            & \cellcolor{red!40}25.10 & \cellcolor{red!40}0.834 & \cellcolor{red!40}0.155 & 286.72 \\
Flash3D~\cite{szymanowiczFlash3dFeedforwardGeneralisable2025} + \ours{} (IM)            & \cellcolor{red!20}25.04 & \cellcolor{red!20}0.826 &\cellcolor{red!40}0.155 & \cellcolor{red!40}143.36 \\
EcoSplat~\cite{parkEcoSplatEfficiencycontrollableFeedforward2026a} & 15.98 & 0.528 & 0.425 & 294.91\\
F4Splat~\cite{kimF4SplatFeedForwardPredictive2026a} & 22.84 & 0.764 & \cellcolor{red!20}0.184 & \cellcolor{red!20}147.46 \\
\bottomrule
\end{tabular}%
}
\end{table}

\begin{figure*}[t] 
  \centering
  \includegraphics[width=\linewidth]{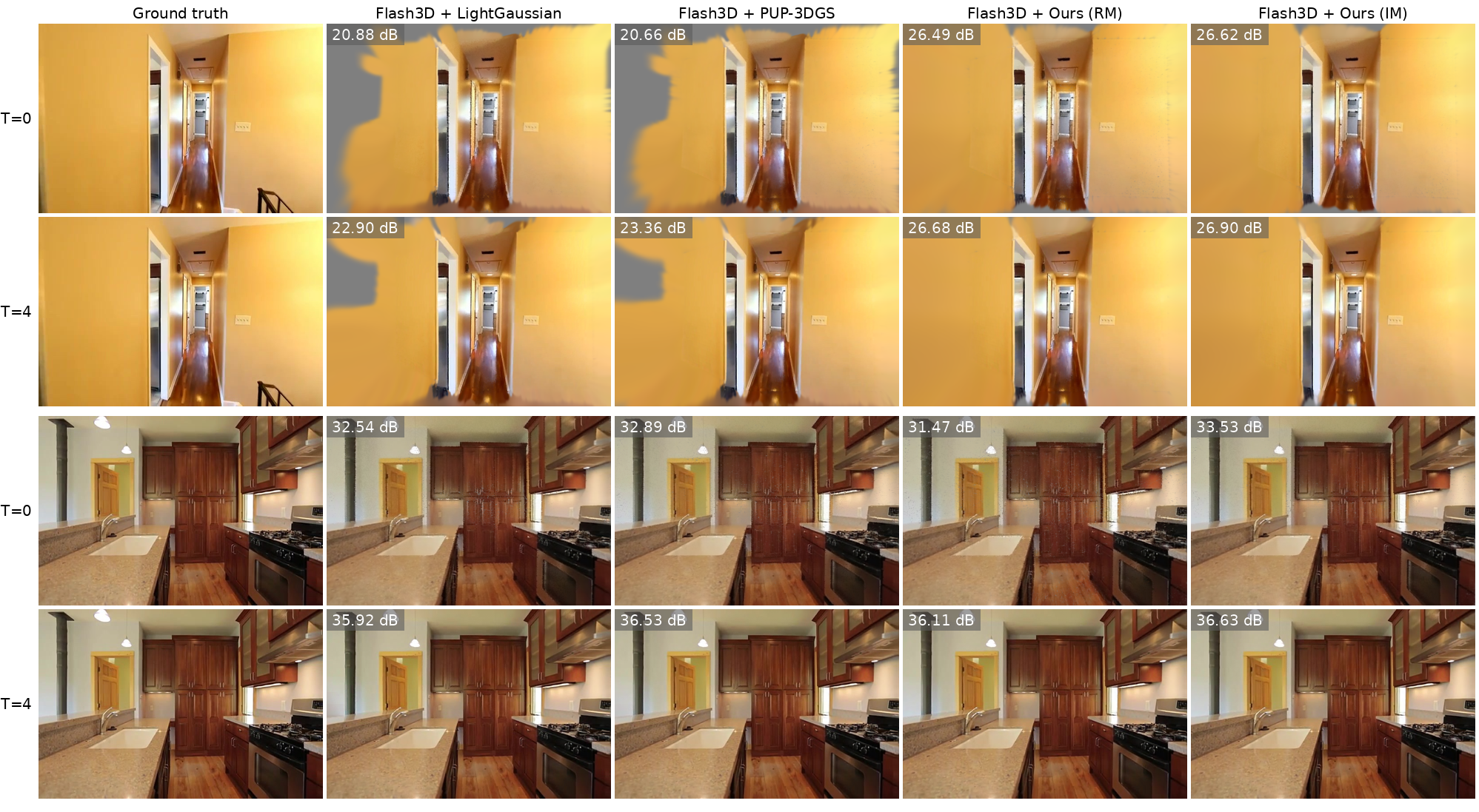}
  \caption{Visual comparison of Flash3D~\cite{szymanowiczFlash3dFeedforwardGeneralisable2025}, Flash3D + \ours{} (RM, IM), and other post-hoc methods such as LightGaussian~\cite{fan2024lightgaussian} and PUP 3D-GS~\cite{hanson2025pup}, applied to Flash3D. Flash3D\cite{szymanowiczFlash3dFeedforwardGeneralisable2025} is shown without pruning as a reference; the other methods use the same keep ratio $r=0.5$. We compare both without ($T=0$) and with ($T=4$) recurrent refinement.}
  \label{fig:compare_posthoc}
\end{figure*}

\subsection{Comparison with Post-Hoc Pruning Methods}
\label{sec: post-hoc}
In Section~\ref{sec: comp-ana}, we compared the pure effect of our importance-score-based pruning method against other post-hoc pruning methods. To verify that the conclusion is not specific to the unrefined setting, we repeat the comparison with recurrent refinement. For each criterion, we train a separate refinement module from scratch, replacing only the importance score and keeping the pruning stage, module architecture, and training schedule identical; LightGaussian~\cite{fan2024lightgaussian} and PUP 3D-GS~\cite{hanson2025pup} use the same sampling variant as in Section~\ref{sec: comp-ana}. 

As shown in Table~\ref{tab:pruning_rrm}, Importance Mask (IM) remains the best on all metrics. Refinement narrows the margins from $0.31$ dB to $0.19$ dB over Lightgaussian and from $0.24$ dB to $0.12$ dB over PUP 3D-GS, with random selection showing the greatest improvement. These results indicate that refinement partially compensates for less effective pruning criteria, while IM consistently retains its advantage. This relation can be seen also in the qualitative results (Figure~\ref{fig:compare_posthoc}). 

\subsection{Additional In-Domain Visualizations}
\begin{figure*}[p]
\centering
\includegraphics[width=\textwidth]{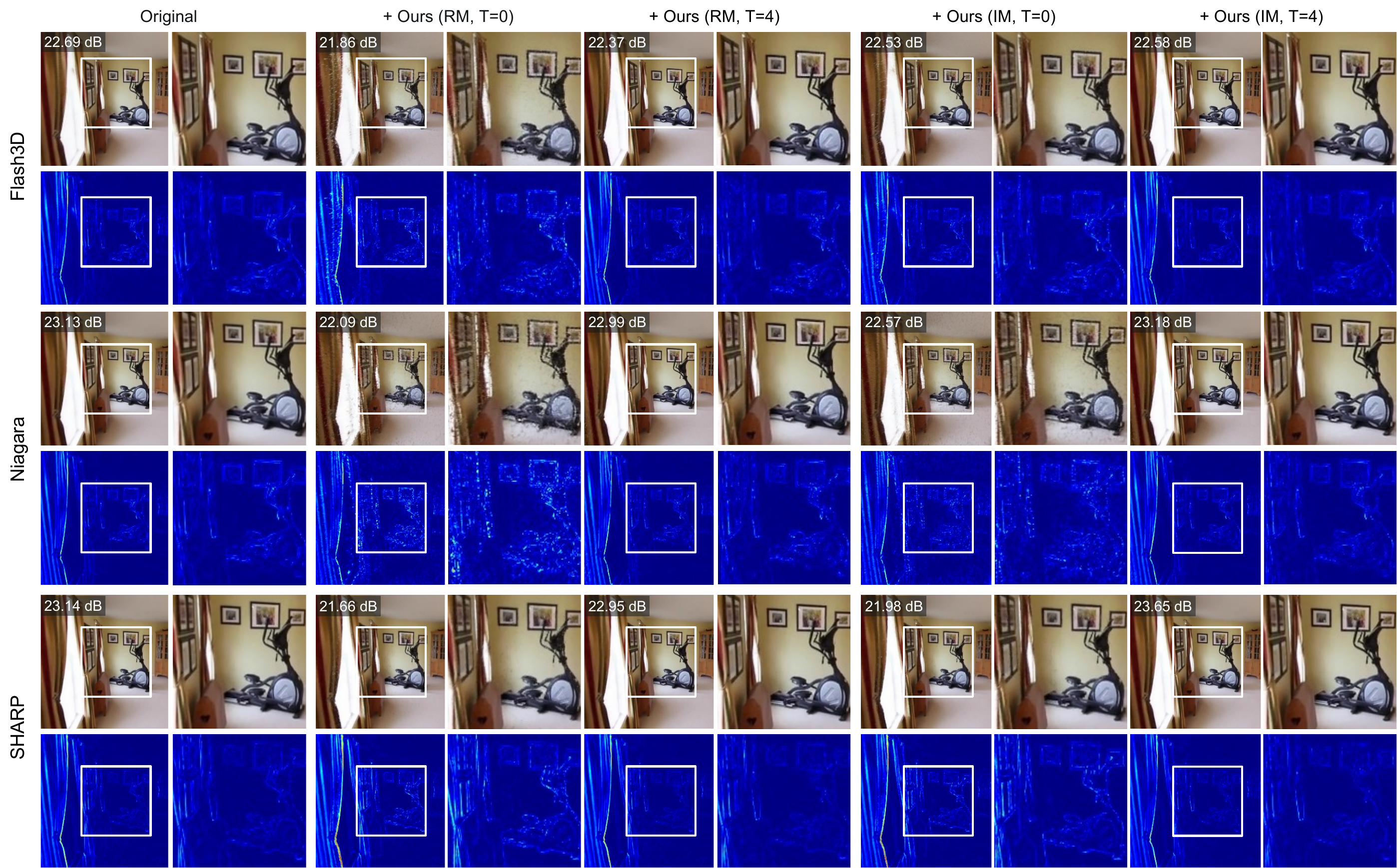}
\caption{Visualizing the rendered target views and corresponding error maps across unrolling iterations on the Flash3D~\cite{szymanowiczFlash3dFeedforwardGeneralisable2025}, Niagara~\cite{wuNiagaraNormalIntegratedGeometric2025a}, and SHARP~\cite{meschederSharpMonocularView2026} backbones ($r=0.5$, Dataset: RealEstate10K~\cite{zhouStereoMagnificationLearning2018c}, Target: $\mathcal{U}[-30, 30]$ frames).}
\label{fig:suppl_compare}
\end{figure*}

Here, we provide extensive qualitative results for three different backbones on the RealEstate10K~\cite{zhouStereoMagnificationLearning2018c} dataset (\figurename ~\ref{fig:suppl_compare}). IM shows low degradation than RM after direct pruning ($T=0$), and our recurrent refinement module improves the rendering quality across all methods. 

\begin{table}[t]
\centering
\footnotesize
\caption{\textbf{Pruning criteria under recurrent refinement.} Each row replaces
only the importance score; the pruning stage, the recurrent refinement module and
its training schedule are identical throughout ($r=0.5$, $T=4$, Dataset: RealEstate10K~\cite{zhouStereoMagnificationLearning2018c}). The first row is the unpruned base model. The \colorbox{red!40}{best} and \colorbox{red!20}{second-best} results among the pruned rows are highlighted.}
\label{tab:pruning_rrm}
\resizebox{\columnwidth}{!}{%
\begin{tabular}{l cccc}
\toprule
Importance score & PSNR $\uparrow$ & SSIM $\uparrow$ & LPIPS $\downarrow$ & \#G (K) $\downarrow$ \\
\midrule
Flash3D~\cite{szymanowiczFlash3dFeedforwardGeneralisable2025}     & \cellcolor{red!40}25.10 & \cellcolor{red!40}0.834 & \cellcolor{red!40}0.155 & \cellcolor{red!20}286.72 \\
LightGaussian~\cite{fan2024lightgaussian}           & 24.85 & 0.822 & 0.161 & \cellcolor{red!40}143.36 \\
PUP 3D-GS~\cite{hanson2025pup}               & 24.92 & 0.824 & \cellcolor{red!20}0.158 & \cellcolor{red!40}143.36 \\
RM             & 24.86 & 0.825 & 0.159 & \cellcolor{red!40}143.36 \\
IM               & \cellcolor{red!20}25.04 & \cellcolor{red!20}0.826 & \cellcolor{red!40}0.155 & \cellcolor{red!40}143.36 \\
\bottomrule
\end{tabular}
}
\end{table}

\subsection{Additional Cross-Domain Visualizations}
Here we provide extensive qualitative results on DL3DV and KITTI to further compare the original Flash3D baseline, RM and IM. Each row corresponds to one test scene. The first column shows the ground-truth target view, followed by predictions from Flash3D, RM, and IM. The numbers in the upper-left corner of each predicted image denote the per-image PSNR computed against the corresponding ground-truth target view.

\begin{figure*}[p]
    \centering
    \includegraphics[width=\textwidth]{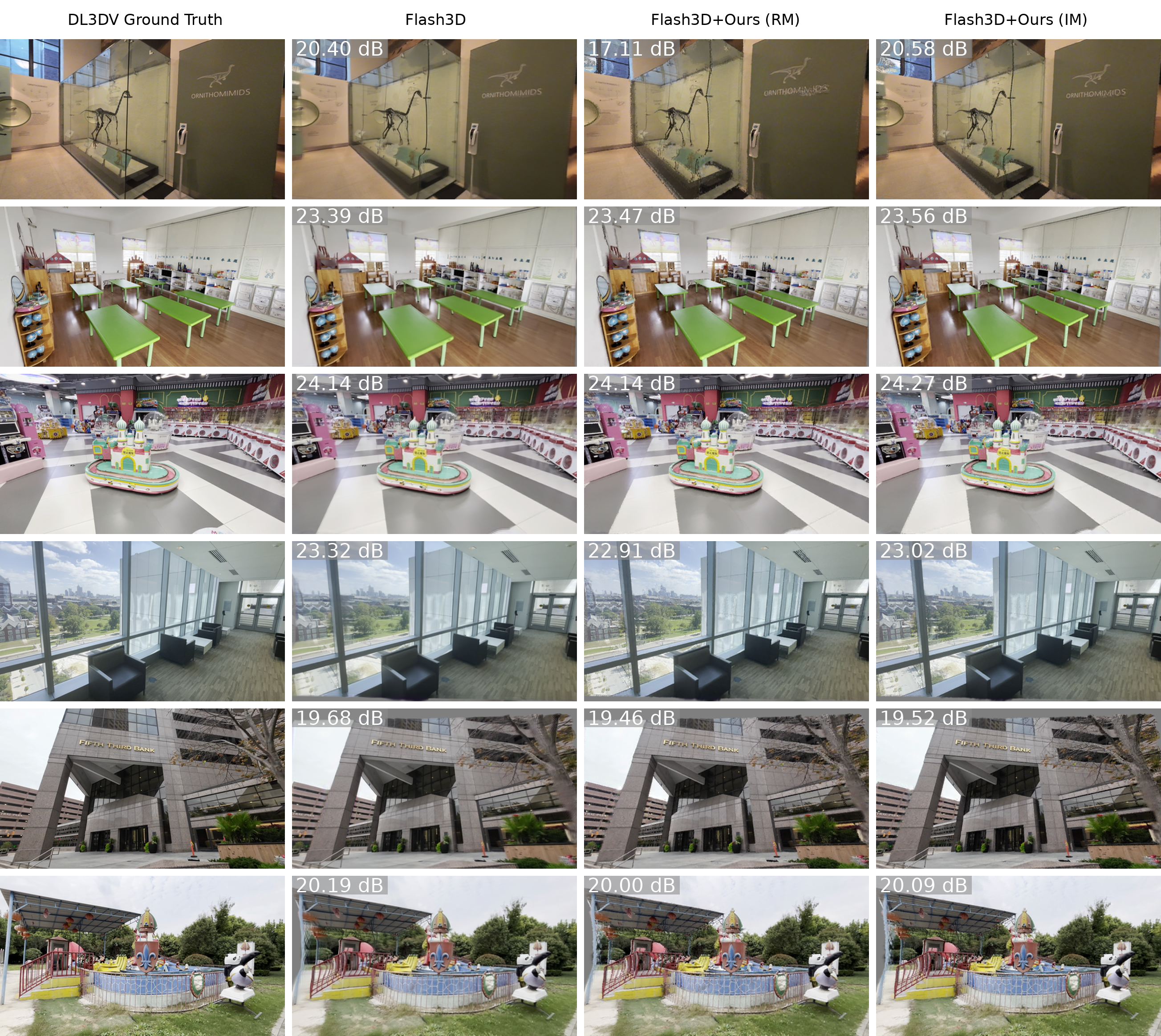}
    \caption{
    Additional qualitative comparison on DL3DV~\cite{ling2024dl3dv}.
    Each row shows one test example with four columns: ground truth, Flash3D~\cite{szymanowiczFlash3dFeedforwardGeneralisable2025}, RM, and IM.
    The PSNR values are computed per image against the ground-truth target view.
    IM obtains slightly higher PSNR than RM in most shown examples, although the perceptual differences are often modest.
    }
    \label{fig:appendix_dl3dv_qualitative}
\end{figure*}

\begin{figure*}[p]
    \centering
    \includegraphics[width=\textwidth]{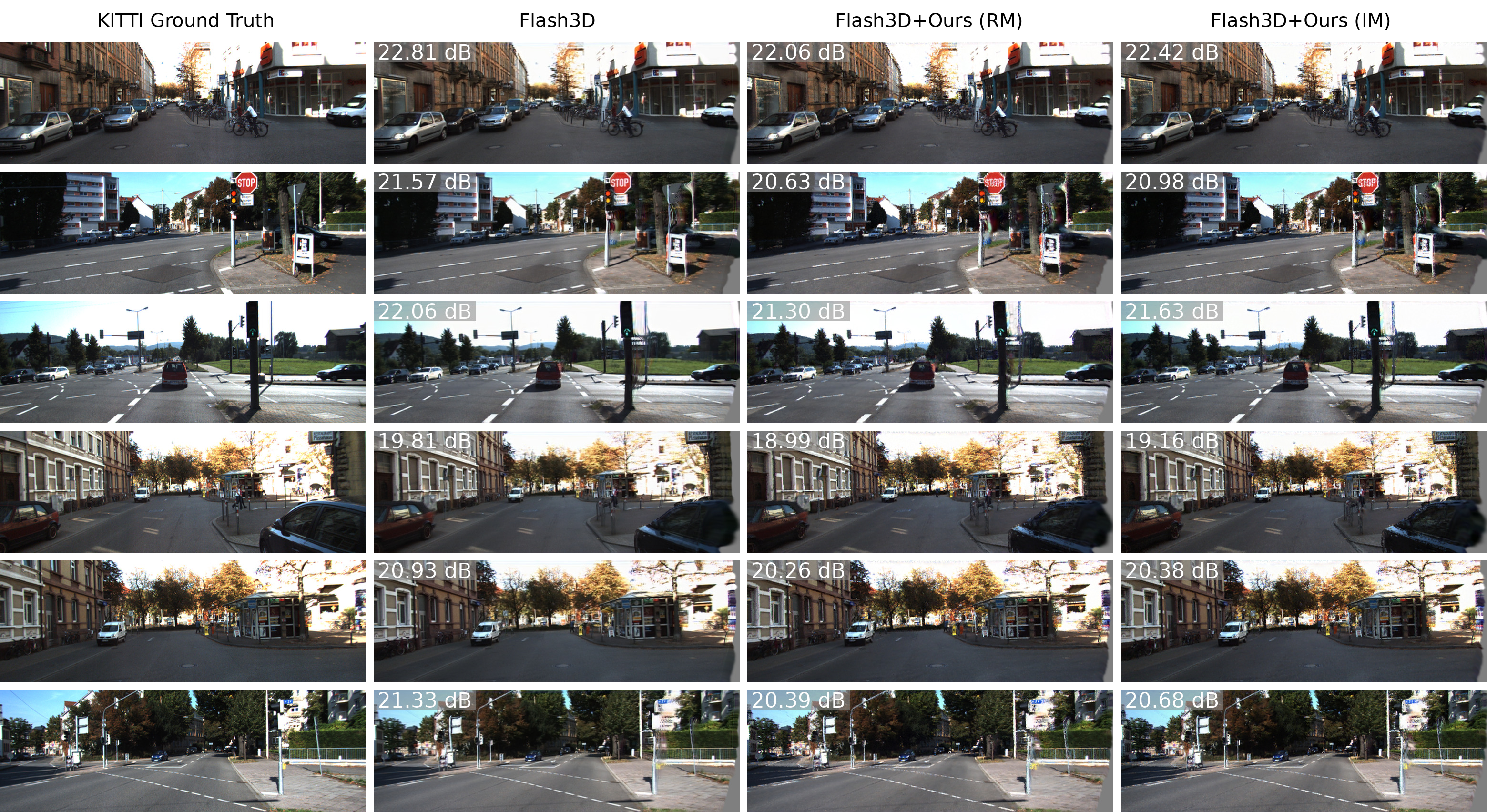}
    \caption{
    Additional qualitative comparison on KITTI~\cite{Geiger2012CVPR, Geiger2013IJRR}.
    Each row shows one stereo target-view prediction with four columns: ground truth, Flash3D~\cite{szymanowiczFlash3dFeedforwardGeneralisable2025}, RM, and IM.
    The PSNR values are computed per image against the ground-truth target view.
    IM is generally comparable to RM visually and achieves slightly higher PSNR on the selected examples.
    }
    \label{fig:appendix_kitti_qualitative}
\end{figure*}

As shown in Figure~\ref{fig:appendix_dl3dv_qualitative} and Figure~\ref{fig:appendix_kitti_qualitative}, the visual differences between RM and IM are often subtle. Nevertheless, IM generally achieves slightly higher per-image PSNR than RM on the selected examples. This suggests that the importance-score-based pruning mechanism provides a more favorable refinement signal, even on challenging datasets.

\end{document}